\documentclass{article}

\usepackage{microtype}
\usepackage{graphicx}
\usepackage{subfigure}
\usepackage{booktabs} % for professional tables
\usepackage{pgfplots,pgfmath,pgffor}
\PassOptionsToPackage{hyphens}{url}\usepackage{hyperref}
\usepackage[accepted]{icml2026}

\usepackage{amsmath}
\usepackage{amssymb}
\usepackage{mathtools}
\usepackage{amsthm}
\usepackage{ dsfont }
\usepackage{mathrsfs}

\usepackage[capitalize,noabbrev]{cleveref}

\theoremstyle{plain}

\theoremstyle{definition}
\newtheorem{definition}{Definition}[section]

\theoremstyle{definition}
\newtheorem{example}{Example}[section]
\newtheorem{remark}{Remark}[section]
\newtheorem{claim}{Claim}[section]
\newtheorem{altdef}{Alternative Definition}[section]
\newtheorem{objection}{Objection}[subsection] % counter resets every \subsection
\usepackage[textsize=tiny]{todonotes}

\usepackage[toc,page,header]{appendix}
\usepackage{minitoc}
\usepackage[x11names,dvipsnames]{xcolor} 

\definecolor{glaucous}{rgb}{0.38, 0.51, 0.71}
\hypersetup{
    colorlinks=true, 
    linkcolor=glaucous,
    filecolor=Blue,      
    urlcolor=Blue,
    citecolor=Blue % For biblatex.
    }

\definecolor{process}{HTML}{1874CD}
\definecolor{novel}{HTML}{D2691E}
\definecolor{rules}{HTML}{009E60}
\definecolor{beliefs}{HTML}{C71585} % I was finding the light pink (#e68fac) a little hard to read 
\definecolor{evidence}{HTML}{9932CC} %{986960}
\definecolor{general}{HTML}{B03060}
\definecolor{conclusions}{HTML}{B22222} %{734F96}{990000}
\definecolor{goal}{HTML}{a0740a}

\usepackage{lipsum}

\usepackage{enumitem}
\setlist[itemize]{noitemsep, topsep=0pt}
\setlist[enumerate]{noitemsep, topsep=0pt}

\usepackage{adjustbox}
\usepackage{tabularray}
\UseTblrLibrary{booktabs}
\usepackage{array}

\usepackage{amsmath}
\usepackage{amssymb}
\usepackage{mathtools}
\usepackage{amsthm}
\usepackage{dsfont} % mathbb for numerals
\usepackage{bbm}    % mathbb for numerals

\DeclareMathOperator*{\argmax}{arg\,max}

\usepackage{tikz}
\usepackage{tikz-3dplot} % glow on edges
\usetikzlibrary {arrows.meta,arrows,patterns,decorations.pathreplacing,chains,positioning,shapes.geometric,shapes.callouts,calc,backgrounds,trees,pgfplots.fillbetween,quotes,calligraphy,automata}

\usepackage{ragged2e}

\usepackage{soul}

\definecolor{working_def}{HTML}{C9C0BB}
\definecolor{position}{HTML}{E6E6FA}

\usepackage[most]{tcolorbox}
\AtBeginEnvironment{tcolorbox}{\footnotesize}
\newenvironment{position}
    {\begin{tcolorbox}[enhanced,%breakable,
    attach boxed title to top center={yshift=-3mm,yshifttext=-1mm},
    colback=position!5,colframe=position,colbacktitle=white,boxrule=1pt,coltext=black,%colframe=black!50
    title=Core Positions,
    fonttitle=\color{black}\scshape,%\itshape,%fontupper=\ttfamily,
    boxed title style={size=small,colframe=position,colback=position!5,boxrule=1pt} ]
    \vspace{2mm}
    }
    { 
    \end{tcolorbox}
    }
\newenvironment{workingdef}
    {\begin{tcolorbox}[enhanced,%breakable,
    attach boxed title to top center={yshift=-3mm,yshifttext=-1mm},
    colback=working_def!5,colframe=working_def,colbacktitle=white,boxrule=1pt,coltext=black,%colframe=black!50
    title=Reasoning --- Informal Definition,
    fonttitle=\color{black}\scshape,%\itshape,%fontupper=\ttfamily,
    boxed title style={size=small,colframe=working_def,colback=working_def!5,boxrule=1pt} ]
    \vspace{2mm}
    }
    { 
    \end{tcolorbox}
    }
    { 
    \end{tcolorbox}
    }

\newenvironment{fullchecklist}
    {\begin{tcolorbox}[enhanced,%breakable,
    attach boxed title to top center={yshift=-3mm,yshifttext=-1mm},
    colback=position!5,colframe=position,colbacktitle=white,boxrule=1pt,coltext=black,%colframe=black!50
    title=Reasoning Research Checklist,
    fonttitle=\color{black}\scshape,%\itshape,%fontupper=\ttfamily,
    boxed title style={colframe=position,colback=position!5,boxrule=1pt} ]
    \vspace{2mm}
    }
    { 
    \end{tcolorbox}
    }

\newif\ifcomments
\commentstrue

\newcommand{\papertitle}[0]{Reasoning is a Learnable Rule-Based Process} 

\icmltitlerunning{Position: \papertitle}

\begin{document}

\twocolumn[

\icmltitle{Position: \papertitle}



\icmlsetsymbol{equal}{*}
%\icmlsetsymbol{msr}{\textdagger}
%\icmlsetsymbol{cornell}{\textdaggerdbl}

\begin{icmlauthorlist}
\icmlauthor{Rachel Lawrence}{msr,equal}
\icmlauthor{Jacqueline Maasch}{cornell,equal}
\end{icmlauthorlist}

\icmlaffiliation{msr}{Microsoft Research, Cambridge, UK}
\icmlaffiliation{cornell}{Cornell Tech, Department of Computer Science, New York, NY}

\icmlcorrespondingauthor{Rachel Lawrence}{rachel.lawrence@microsoft.com}
\icmlcorrespondingauthor{Jacqueline Maasch}{jam887@cornell.edu}

\icmlkeywords{Machine Learning, ICML, reasoning, trustworthy}

\vskip 0.3in
]



\printAffiliationsAndNotice{\icmlEqualContribution} % otherwise use the standard text.

% FOR TOC IN APPENDIX
\doparttoc % Tell to minitoc to generate a toc for the parts
\faketableofcontents % Run a fake tableofcontents command for the partocs
%\part{} % Start the document part
%\parttoc % Insert the document TOC

\begin{abstract}
  Autonomous reasoning is among the most scientifically and economically motivating topics in AI today. Historically the purview of symbolic AI, recent advances have mainly emerged from deep probabilistic generative models. Despite immense interest and rapid progress, the generative AI community has not clearly converged on operational definitions for reasoning and often implicitly rejects the historical treatment of this topic in logic and verifiable automated reasoning. This position contends that definitional ambiguity leaves the construct validity of reasoning evaluation unverifiable, undermining quantifiable progress toward trustworthy autonomous reasoning. We also contend that this ambiguity is addressable. To that end, we provide (1) operational definitions based on a synthesis of the literature, positioning valid and sound reasoning as a \textit{learnable rule-based process}; and (2) a checklist for best practices in the communication of AI reasoning research.
\end{abstract}

\section{Introduction} \label{sec:intro}

% TODO: reduce to minimal set of major subtheses, link each to the \S section that we will write about them (i.e., each subthesis should correspond to a subsection)
\begin{figure}[!t]
    \centering
    %%%%%%%%%%%%%%%%%%%%%%%%%%%%%%%%%
    %%%%%%%%%%%%%%%%%%%%%%%%%%%%%%%%%
    %%%%%%%%%%%%%%%%%%%%%%%%%%%%%%%%%
    \begin{workingdef}
    {\footnotesize
        \textit{The process of selecting and applying sequences of rules that act on prior beliefs and current evidence to obtain principled belief updates in evolving states.}
    }
    \end{workingdef}
    %%%%%%%%%%%%%%%%%%%%%%%%%%%%%%%%%
    %%%%%%%%%%%%%%%%%%%%%%%%%%%%%%%%%
    %%%%%%%%%%%%%%%%%%%%%%%%%%%%%%%%%
    \begin{position}
    {\footnotesize
        \begin{enumerate}[
        wide, 
        labelindent=0pt,
        itemsep=1ex,
        label=\textbf{Thesis \arabic*},
        ref={Thesis \arabic*}
        ]
            \item \label{thesis:definition} \textbf{\textit{Define, then  measure}. }  Operational definitions should be stated for the reasoning phenomena under investigation, and the construct validity of reasoning evaluation should be justified with respect to these definitions.
            \item \label{thesis:rules} \textbf{\textit{Reasoning is a learnable rule-based process}.} Reasoning is a process of exact rule application, not an output. Learnable rules map reasoning inputs to outputs and can encompass theorems, functions, policies, etc., including rules pertaining to stochasticity, uncertainty, and approximation.
            \item \label{thesis:validity} \textbf{\textit{Rule-based reasoning is valid}.}  The \textit{validity} of a reasoning process arises from exact rule application, independent of rule selection. 
        \end{enumerate}
    }
    \end{position}
    %\vspace{-3mm}
    \caption{Core theses of this position.}
    \label{fig:position}
\end{figure}

The prospect of AI reasoning is among the most scientifically and economically motivating advancements of the current era. Recent progress  has been fueled by the remarkable empirical performance of large reasoning models (LRMs): large language models (LLMs) fine-tuned for \textit{reasoning tasks} (Glossary  \ref{def:reasoning_task}; \citealt{huang2023towards}). A wave of benchmarking successes invites many questions: Is autonomous reasoning an emergent behavior that arises with scale  \citep{wei2022emergent,gonzalez2024does}? Is it a foregone conclusion that LRMs can be formally characterized as autonomous reasoners? The answers are contingent on how reasoning is defined. %Building on Feynman's adage,\footnote{\textit{What I cannot \underline{create}, I do not understand.}} we are motivated by the following: \textit{what I cannot define, I do not understand.} 

So then, \textit{what is reasoning}? 
Though a universal definition may not exist, we argue that practical operational definitions (Glossary \ref{def:operational}) are achievable but not yet in widespread use in generative AI. 
%\textbf{This work advocates for the use of formal operational definitions for reasoning based on \textit{validity},\footnote{See Definition \ref{def:reasoning_formal} (\textit{Reasoning}) and Definition \ref{def:validity} (\textit{Validity}).} positioning reasoning as a learnable process grounded in exact rule application (Fig. \ref{fig:position}).} \jm{Note that we use the term as in the logic community.}
\textbf{This work advocates for the use of formal operational definitions for reasoning, and positions valid and sound reasoning as a learnable process grounded in exact rule application (Fig. \ref{fig:position}).}

\textbf{Lack of Consensus}\; Reasoning remains an elusive target in AI, despite prolific study across the history of human thought (Appendix \ref{sec:historical}). Though claims of emergent reasoning in generative AI are now commonplace, ``there is not a clear definition of what it entails'' \citep{huang2023towards}. In the absence of consensus on what formally constitutes reasoning in generative AI, we observe a normalization of research outputs that claim to study, improve, measure, or promote AI reasoning without rigorously defining the form of reasoning under investigation. This definitional void enables shifting goalposts and leaves the construct validity of reasoning evaluation unverifiable, obscuring clear progress toward human-level reasoning. Avoidance of formal definitions may owe to an implicit assumption that reasoning is an intuitive concept requiring no explicit definition; evasion of the hard work of devising operational definitions; silent rejection of historical definitions from symbolic AI; or (un)intentional conflation of benchmark accuracy with reasoning itself. We aim to make the risks of such avoidance evident, and to suggest alternative paths. Namely, we do not see a justification for \textit{reinventing reasoning} in the context of generative AI: we project that operational definitions that are \textit{method agnostic} -- simultaneously compatible with symbolic, neural, and hybrid methods -- will provide greater conceptual unification and research value in the long run.

\textbf{Reasoning Zombies \& Other Hard Problems} \;
The black-box design and natural language interface of LRMs  present a nontrivial challenge: differentiating true reasoning from reasoning-like speech. The latter represents superficial emulation: \textit{talking like a reasoner} with no guarantees that conclusions arose from anything more than memorization, guessing, Clever Hans effects \citep{lapuschkin2019unmasking,kauffmann2025explainable}, %Gettier cases \cite{gettier1963justified,zagzebski1994inescapability}, 
or some other man-behind-the-curtain \citep{mitchell2025artificial}. This challenge is not unique to AI reasoning: parallels can be drawn to human cognitive testing and to distinguishing intelligence, understanding, and intentionality from sophisticated emulation, as canonized by the Turing Test \citep{turing1950test,pinar2000turing} and the Chinese room argument \citep{searle1980minds,searle1990brain,dennett1980milk,hauser1997searle}. This evokes a rough analogue of the philosophical zombie (\textit{p}-zombie) thought experiment, which we term the \textit{reasoning zombie} (\textit{r}-zombie). In the classic thought experiment, \textit{p}-zombies are systems that superficially behave like conscious beings, yet lack any conscious internal experience \citep{chalmers1997conscious,chalmers2020spatiotemporal}. 
Analogously, \textit{r}-zombies are systems that superficially behave as autonomous reasoners, but lack valid
internal reasoning mechanisms. \textit{Perfect} \textit{r}-zombies, which behave identically to true reasoners in all circumstances, remain purely theoretical. However, we argue that (1) \textit{imperfect} AI \textit{r}-zombies have already come into existence; (2) differentiating AI \textit{r}-zombies from AI reasoners is theoretically and, often, empirically possible; and (3) we must carefully determine when real-world use cases \textit{require reasoners}, and when \textit{r}-zombies suffice.

\textbf{What This Position Is \;} Our core theses (Fig. \ref{fig:position}) follow from two main problems.
\begin{enumerate}[label=\textbf{P\arabic*},ref=P\arabic*]
    \item \label{problem:p1} Reasoning in generative AI has experienced unnecessary and addressable definitional ambiguity, where imprecise and overloaded definitions are often misaligned with historical treatments of this topic in AI and philosophy (when definitions are provided at all).
    \item \label{problem:p2} This breeds mismeasurement, promotes an illusion of shared understanding among researchers, and subverts measurable progress toward trustworthy AI reasoning. 
\end{enumerate}

\textbf{What This Position Is Not \;} We do not claim that the AI community must converge on one universal definition for reasoning. We do not attempt to formally characterize the kind or extent of reasoning that LRMs can perform, nor do we propose practical implementations for improving LRM reasoning. This position is not an endorsement for or against symbolic AI, purely data-driven approaches, nor neuro-symbolic AI.  We do not make claims about reasoning in natural intelligences, nor do we argue that reasoning implies understanding or consciousness. %\jm{Not sure where to put this comment from the rebuttal.} \rebuttal{Questions of cognition, while philosophically significant, are orthogonal to our goal of formalizing automatable reasoning. We do not take a stance on whether syntactic manipulation constitutes cognition and we explicitly decouple the mechanistic exact reasoning process from the reasoner performing it.}

\textbf{Contributions \& Artifacts} 
\begin{enumerate}
    \item \textbf{An operational definition for reasoning as a learnable, rule-governed process.} Based on a synthesis of the literature, we introduce an  operational definition for AI reasoning for general use and community discussion (\S \ref{sec:definitions}). Per \ref{thesis:definition}, we express this definition in (1) natural language for intuition; (2) mathematical notation for concretization; and (3) pseudocode (Algorithm \ref{alg:reasoning}). Operationalization is illustrated by trivial Python implementations.\footnote{\href{https://github.com/jmaasch/valid\_reasoning}{https://github.com/jmaasch/valid\_reasoning}}  
    We apply our definitions to  special cases, including logical deduction, Bayesian inference, reinforcement learning (RL), and probabilistic next token prediction. In \S\ref{sec:alternatives}, we address rebuttals to our definitions and theses.
    \item \textbf{Recommendations for scientific communication.} We propose a checklist of community guidelines that complies with \ref{thesis:definition}–\ref{thesis:validity} (Appendix \ref{sec:checklist}).
\end{enumerate}

\subsection{Problem Significance: Why Do We Care?} 
\label{sec:problem_significance}

The import of \ref{problem:p1} and \ref{problem:p2} lies primarily in the following: (1) reasoning is a necessary (but not sufficient) precondition for artificial general intelligence (AGI); (2) AI evaluation faces a construct validity crisis, which has spilled over into reasoning evaluation; and (3) the rate of user uptake for LRMs has outpaced evidence of trustworthy reasoning.

\textbf{Reasoning is a Precondition for AGI} \;  \label{sec:agi}  Though contentious, AGI is widely viewed as a north star for contemporary AI research \citep{morris2024position,blili2025stop}. However, lack of community consensus on the definition and measurement of AGI hinders progress. A recent effort to operationalize AGI promotes a taxonomy of subcomponents and benchmark-based means of measuring these \citep{hendrycks2025definition}. Based on human cognitive testing, this taxonomy emphasizes \textit{on-the-fly reasoning} as an essential component of measurable AGI. We agree with \citet{hendrycks2025definition} that the ability to reason is a necessary (but not sufficient) precondition for AGI. An excess of valuable use cases aside, this alone is sufficient to justify AI reasoning as a critical area of inquiry. However, like AGI, a shroud of ambiguity, confusion, and debate looms over the definition and measurement of AI reasoning. If reasoning is a necessary precondition for AGI, then measurable progress toward clearly defined reasoning will be necessary for measurable progress toward AGI. %For example, we could pose the open question: can an \textit{r}-zombie, perfect or imperfect, ever achieve AGI? Whether the answer is positive or negative, answering such questions will require theory and methods for differentiating AI reasoners from \textit{r}-zombies.

\textbf{Construct Validity is Underemphasized} \; \label{sec:construct_validity} 
Recent waves of generative AI tend to emphasize exploratory research and empirical evaluations over hypothesis-driven confirmatory research, proof of theoretical guarantees, or formal verification (Glossary \ref{def:formal_verification}; \citealt{herrmann2024position}). Historically, empirical fields have taken precautions against mismeasurement via \textit{construct validation} (Glossary \ref{def:construct_validity}; \citealt{cronbach1955construct}): justifying that experimental measures capture the phenomena of interest by devising operational definitions that relate latent abstract constructs 
(e.g., intelligence, bias, ideology) to measurable proxies. And yet, ``validity and other quality criteria of empirical research have gained little attention in ML so far'' \citep{herrmann2024position}, eliciting commentary that evaluation in natural language understanding is largely ``broken'' \citep{bowman2021will} and that AI evaluation must ``mature into a proper `science''' \citep{weidinger2025toward}. Benchmarking with static datasets is the standard framework for generative AI evaluation, but it faces multiple crises, e.g.: poor construct validity \citep{wallach2025position,alaa2025position}, data contamination \citep{white2025livebench}, overfitting, minimal quality control, gaming, SOTA hacking, and selective reporting  \citep{cheng2025benchmarking}. We observe several points of risk for construct validity in current reasoning evaluation strategies, including: 
\begin{enumerate}[leftmargin=*]
    \item \textit{A process and its product should not be conflated.} Reasoning benchmarks often treat question-answering (QA) accuracy as a proxy for reasoning  (\citealt{clark2018think}, \textit{inter alia}). However, final-answer accuracy does not guarantee the mechanism by which the answer was generated \citep{zhang2025dagmath}, and we echo \citet{chollet2019measure} and \citet{simon2000bounded} on the risks of conflating a process with its artifacts.\footnote{We echo \citet{chollet2019measure} on the risks of ``confusing the process of intelligence'' (reasoning, in our case) ``with the artifact produced by this process'' (e.g., QA responses), ignoring the generating mechanism: ``In the case of AI, the focus on achieving task-specific performance while placing no conditions on \textit{how the system arrives at this performance} has led to systems that, despite performing the target tasks well, \textit{largely do not feature the sort of human intelligence that the field of AI set out to build}'' (original emphasis). \citet{simon2000bounded} similarly argued that a theory of bounded rationality ``will be as much concerned with [...] the quality of the processes of decision, as with [...] the quality of the outcome.''} We contend that (i) reasoning is a \textit{process} and not an \textit{output} (Figure \ref{fig:position}), (ii) accurate QA final-answers can be obtained by \textit{r}-zombies via non-reasoning behaviors, and thus (iii) accurate QA is not sufficient for demonstrating that reasoning has taken place.
    \item \textit{Chain-of-thought (CoT) traces are not trustworthy explanations.}  If intermediate reasoning steps are evaluated, CoT ``reasoning traces'' often serve as a stand-in for the LRM's internal reasoning process. However, CoT is neither necessary nor sufficient for obtaining trustworthy explanations \citep{barez2025chain}. Though attractively anthropomorphic, CoT is not guaranteed to be faithful to the model's internal decision-making    \citep{turpin2023language,lyu2023faithful,lanham2023measuring,kambhampati2025stop,zhang2025dagmath}. Mid-CoT shifts (e.g., ``aha!'' moments) may be rarer and less impactful than previously thought, reflecting unstable inference rather than true self-corrective reasoning \citep{daliberti2026aha}. We contend that an imperfect \textit{r}-zombie could produce convincing but untrustworthy (or adversarial) CoT by emulating reasoning structure rather than content \citep{li2025structure_not_content}.
    \item \textit{Evaluations should disentangle reasoning from recall.} Many reasoning and intelligence benchmarks are easily gamed by instilling near-unlimited priors and experience through large-scale pre- and post-training \citep{chollet2019measure}. This is a core challenge in differentiating reasoning from recall in LRMs \citep{huyuk2025,xu2025reimagine,maasch2025ccr}, raising the potential for \textit{r}-zombies that lack %out-of-distribution 
    robust reasoning mechanisms yet are SOTA on benchmarks. Evidence of this potential can be found in the fragility of benchmark performance under superficial perturbations, such as reworded premises, altered numerical values or variable names, or the injection of irrelevant details  \citep{shojaee2025illusion,mirzadeh2025gsm,xu2025reimagine}. 
\end{enumerate}

\textbf{Usership Outpaces Trustworthiness} \label{sec:trust_usership} \; 
Science is fundamentally a ``collective epistemic enterprise,'' and as such, \textit{epistemic trust} (Glossary \ref{def:epistemic_trust}) underpins scientific integrity \citep{wilholt2013epistemic}. 
Epistemic trust in machine reasoning has been championed most in mathematical domains, as epitomized by the Lean language for automated theorem proving \citep{de2015lean}. Lean addresses the ``trust bottleneck'' through formal verification, providing guarantees on correctness \citep{castelvecchi2023will}. However, the shift from deterministic systems and formal verification to probabilistic generative AI has raised new specters for epistemic trust \citep{song2026failures}, including evidence that hallucination is a feature and not a bug \citep{xu2024hallucination, bastounis2024consistent}, accuracy collapse as task complexity scales \citep{shojaee2025illusion}, poor out-of-distribution generalization  \citep{chollet2024arc,mirzadeh2025gsm,xu2025reimagine}, and low explainability. LLM-hallucinated citations \citep{gptzero2025citations,sakai2026hallucitation} and other sources of epistemic distrust in peer review at flagship AI conferences have elicited calls for reform \citep{kim2025position}. Rampant accusations of ``AI hype'' \citep{placani2024anthropomorphism,mit2025hype} coincide with broader linguistic trends: decreased hedging of uncertainty in scientific communication \citep{yao2023promoting} mirrors trends across diverse English text sources \citep{scheffer2021rise}, reflecting a normalization of language that exaggerates confidence and obscures limitations. 
Meanwhile, 59\% of AAAI survey respondents agreed that AI trustworthiness remains ill-defined, while 60\% predicted that neither trustworthiness nor factuality would be solved in the near future \citep{rossi2025aaai}.  See Appendix \ref{appendix:trust} for further discussion.

\section{Operationalizing Valid \& Sound Reasoning}
\label{sec:definitions} 
Defining reasoning is a nontrivial challenge, as represented by millennia of scholarly effort (Appendix \ref{sec:historical}). \citet{broome2013rationality} admits that five years of iterative self-correction were required to reach an understanding of reasoning. Thus, it is unsurprising that researchers can struggle to choose authoritative definitions for use in contemporary AI.

As a step toward addressing \ref{problem:p1} and \ref{problem:p2}, we provide working definitions for reasoning that take a rule-centric perspective while remaining suitable for neural and neuro-symbolic applications (\ref{thesis:definition}, \ref{thesis:rules}). Definitions are a synthesis of prior efforts from diverse domains, including computing, philosophy, and the social sciences. %They are intended to be a starting point for discussion and a community reference for those that require practical, operational definitions. 
We address \textit{reasoning} and \textit{reasoners} in general (\S\ref{sec:working_defs}), domain-specific cases (\S\ref{sec:domain_specific_reasoning}), and  \textit{validity} and \textit{soundness}  (\S\ref{sec:validity}).  

\subsection{Working Definitions for Reasoning}

\label{sec:working_defs}

\subsubsection{Intuition in Natural Language}
\label{sec:informal_defs}

We begin with plain English to establish intuition. Colored terms denote core components, which we judge to be conserved elements from across the historical literature.

\begin{definition}[Reasoning, informal]
\label{def:reasoning_informal}
    The {\color{process}\textit{process of selecting and applying}} {\color{rules}\textit{sequences of rules}} that act on {\color{beliefs}\textit{prior beliefs}} and {\color{evidence}\textit{current evidence}} to  obtain {\color{conclusions}\textit{principled belief updates}} in {\color{novel}\textit{evolving states}}. 
\end{definition}

\begin{definition}[Reasoner, informal] \label{def:reasoner}
    A {\color{goal}\textit{goal-oriented}} decision-maker that implements reasoning. 
\end{definition}
This conceptualization is closely related to arguments by \citet{chollet2019measure} that intelligence is a process and by \citet{broome2013rationality} that reasoning is a process, ``something a person \textit{does}'' (emphasis added), and a ``rule-governed operation'' (p. xii). Reasoning is fundamentally an \textit{epistemic process}: rules are operators whose operands are \textit{information}, which can be partitioned into evidence, beliefs, and other rules. %\rebuttal{See \citet{pearl1990reasoning} for another account of \textit{belief} as informed by \textit{evidence}. See also \citet{hayes1981frame} for discussions of belief.}

Framing reasoning as a sequential process implies a notion of time $t$. We can conceptualize a time-dependent snapshot of the reasoner's internal world representation, which we refer to as the \textit{state} at time $t$.\footnote{Note that the state is not necessarily a \textit{world model} as commonly conceived in RL or structural causal modeling \citep{richens2024robust,richens2025general,maasch2025causalarc}: it is not necessarily predictive of the dynamics governing an evolving environment nor sufficient for causal identifiability. Further, it may be only partially observed or partially stored in memory. }

\begin{definition}[State, informal] \label{def:state}
    The set of all parameters that are pertinent to the reasoner at time $t$, including some subset of the historical record of beliefs, evidence, and rules.
\end{definition}

We provide further intuition for each component of Def. \ref{def:reasoning_informal}.

{\color{process}\textbf{Process}} \; Reasoning is a dynamic process, not an output. Thus, reasoning entails $T \geq 1$ hops, stages, time steps, or \textit{reasoning steps}. 
This process implies a design component: sequences of rules or actions are chosen by the reasoner according to some justification. The process of \textit{selection} is where agency, intelligence, or creativity may come into play, while the process of \textit{execution} necessitates exactness and rigor. Note that it may be perfectly reasonable for the selection criterion to be random selection.

{\color{goal}\textbf{Goals}} \; The reasoner generally executes a reasoning process to achieve some outcome of interest. This outcome is the \textit{goal} one is reasoning toward: the answer to a complex question, the solution to a puzzle, the shortest path through a maze, a mathematical proof, the optimal action to take under resource constraints, etc. In distinguishing the goal-directed reasoner from the reasoning process itself, we highlight that the \textit{validity} of the reasoning process is not necessarily tied to successful attainment of a goal (see \S\ref{sec:validity}). In practice, we can encode the goal in a stopping rule, where reasoning terminates when the rule is satisfied. We do not restrict our notion of goals to the formal sense used in RL \citep{sutton1998reinforcement}, though it is compatible with this interpretation.

%\rebuttal{Rules can be viewed as \textit{operators} whose operands are (1) exogenous or extrinsically obtained information (\textit{evidence}); (2) endogenous or intrinsically generated information (\textit{beliefs}); and/or (3) other rules. Evidence acts as an input to the rule set, while beliefs can be both inputs and outputs. We can model evidence as a continuous stream of received data that is updated at each step $t$ or at intervals. Prior beliefs are the outputs of previous reasoning steps, i.e., intermediate conclusions generated along the reasoning pathway that led to step $t$. Current beliefs denote the conclusions drawn in the transition from $t-1$ to $t$. When $t=T$, current belief is equivalent to the terminal conclusion of the reasoning process.}

{\color{rules}\textbf{Rules}} \;  Collectively, the rule set unambiguously maps the reasoning state at $t-1$ to the state at $t$. Rules can be viewed as operators whose operands are (1) exogenous or extrinsically obtained information (\textit{evidence}); (2) endogenous or intrinsically generated information (\textit{beliefs}); and/or (3) other rules in the rule set (e.g., during rule learning and revision). Evidence acts as an input to the rule set, while beliefs and rules can be inputs and outputs. In general, rules are selected with some justification prior to deployment. Rules can take the form of algorithms, formulae, theorems, axioms, laws, policies, premises, assumptions, decision boundaries, etc. Rules can be extrinsically imposed on the reasoner (i.e., hard-coded by another individual or collective agent, such as a human or government) or they can be learned autonomously from data on-the-fly. Rules can be fixed or continuously updated in light of new information.

{\color{evidence}\textbf{Evidence}} \; Evidence is a form of exogenous or extrinsically obtained information. We can model evidence as a continuous stream of data that is updated at each step $t$ or at intervals. \textit{Current evidence} denotes information presented at $t$,
along with the historical record: aggregated information up to $k \geq 0$ steps prior to $t$. Evidence may be gained directly through sequential interactions with an uncertain environment (as in online RL, field work in the natural sciences, etc.) or provided without direct collection (e.g., retrospective data collected by another agent). In trivial cases, external evidence is the empty set or is provided at $t=0$ and never updated.

{\color{beliefs}\textbf{Prior Beliefs}} \;
While evidence is extrinsically obtained, we model beliefs as a form of endogenous or intrinsically generated information. \textit{Prior beliefs} are the outputs of previous reasoning steps, up to step $t-k$ for $t > k \geq 1$. They are intermediate conclusions along the reasoning pathway that led to step $t$. Often, they are defeasible: they can be overwritten if proven false (e.g., in backtracking proof search), refined if insufficient, or maintained and aggregated with current beliefs at step $t$. They can also be  provided at $t=0$ (e.g., initializing Bayesian priors based on convention when supporting evidence is not yet available).

{\color{conclusions}\textbf{Current Beliefs}} \; 
Current beliefs denote the conclusions drawn in the transition from $t-1$ to $t$. When $t=T$, current belief is equivalent to the \textit{terminal conclusion} of the reasoning process. The nature of the terminal conclusion is a defining property of the type of reasoning performed, e.g.: the output of a function in mathematical reasoning, an optimal action in practical reasoning, a moral verdict in moral reasoning, a judiciary decision in legal reasoning, etc.

{\color{novel}\textbf{Evolving States}} \; A reasoner will generally maintain an \textit{internal representation} of its world state (Def. \ref{def:state}), which updates over time. The existence of an \textit{external environment} is also implied by our choice to model evidence as a stream of extrinsic signals. However, we note that a well-defined concept of external environment is not relevant in all cases (e.g., in some mathematical reasoning domains). Thus, we place no requirements on the existence or direct observability of an external environment, physical world, etc., and only require an internal representation of the world (i.e., the state).  
We use the notion of an \textit{evolving state} broadly to encode all of the above concepts: (1) dynamically updated internal state representations, (2) changing and/or uncertain external worlds, and (3) extrinsic sources of evidence. 

\subsubsection{A Formal Operational Definition}
\label{sec:formal_defs}

Natural language is too ambiguous for measurable definitions in the general case. We offer an operationalization of Def. \ref{def:reasoning_informal} in mathematical notation and pseudocode. Note that Def. \ref{def:reasoning_informal} could admit alternative operational definitions. %We begin by stating our assumptions.

%\begin{enumerate}
%[label=\textbf{A\arabic*},ref=A\arabic*]
%    \item \label{assumption:consciousness} Reasoning is an exact procedure that can be decoupled from consciousness, understanding, and intentionality.
%    \item \label{assumption:automation} Thus, reasoning can be a fully automated process.
%\end{enumerate}
%\rebuttal{Note that we \textit{do not} assume that all components of the reasoner itself can be decoupled and fully automated in the above sense. We leave this much larger open question to future research. Further, the following operationalization is a \textit{mathematical} account that assumes \ref{assumption:consciousness} and \ref{assumption:automation}. We adopt mathematical language because it is familiar and actionable for the theoretical and engineering communities that conduct AI research. Though our assumptions and language may diverge from some alternative accounts in the philosophical literature, we argue that this has practical utility for building real systems. See Objection \ref{objection:consciousness} for competing views.}

\begin{definition}[Reasoning, formal]
\label{def:reasoning_formal}
    Let $\mathcal{S}_t \coloneqq \langle \mathcal{B}_t, \mathcal{E}_t, \mathcal{R}_t \rangle$ denote the reasoner's state at time step $t$, where $\mathcal{B}_t$ denotes current belief, $\mathcal{E}_t$ denotes aggregated evidence up to time $t$, and $\mathcal{R}_t$ denotes the current set of established rules. Then, \textit{reasoning} is the {\color{process}iterated application over  steps} $t$ of {\color{rules}rules} $r \in \mathcal{R}_{t-1}$ to {\color{beliefs}prior beliefs } $\mathcal{B}_{t-1}$ and {\color{evidence}current evidence} $\mathcal{E}_{t}$, by which we obtain {\color{novel}dynamically updated states} $\mathcal{S}_t$, and where every 
    {\color{conclusions}output} $\mathcal{B}_{t}$ for $t>0$ {\color{conclusions} is the result of a rule application} $r(\mathcal{B}_{t-1}, \mathcal{E}_{t})$ to the contents of state $\mathcal{S}_{t-1}$.%{to a reachable state} $\mathcal{S}_{t-k}$ for $t > k \geq 1$. 
\end{definition}

Thus, rules and extrinsic evidence updates are the mechanism by which $\mathcal{S}_t$ changes over time: each $r \in \mathcal{R}_t$ is a function acting on subsets of the current state $\mathcal{S}_t$ to generate some attribute of the next state $\mathcal{S}_{t+1}$. Rule set $\mathcal{R}$, beliefs $\mathcal{B}$, and evidence $\mathcal{E}$ comprising state $\mathcal{S}$ are each elements of a corresponding space $\mathbf{R}$, $\mathbf{B}$, $\mathbf{E}$, and $\mathbf{S}$. $\mathcal{R}$ is a set of functions, with domains and ranges as defined below. Other implementation details, constraints, and type systems defining these spaces are problem-specific.

\begin{definition}[Reasoning components]
\label{def:reasoning_components}
{
% \small
\begin{align*} 
    &t \in [0, ..., T] & \text{Reasoning step.} \\
    &\{ \mathcal{B}_i \}_{i=0}^T,\ \mathcal{B}_i \in \mathbf{B} & \text{Beliefs.} \\
    &\{ \mathcal{E}_i\}_{i=0}^T,\ \mathcal{E}_i \in \mathbf{E} 
    & \text{Evidence.} \\
    &\{\mathcal{R}_i\}_{i=0}^{T},\ \mathcal{R}_i \in \mathbf{R} & \text{Rule set.}\\ 
    &\mathcal{S}_i \coloneqq \langle \mathcal{B}_i, \mathcal{E}_i, \mathcal{R}_i \rangle,\ \mathcal{S}_i\in \mathbf{S} & \text{States.} 
\end{align*} }%
The rule set is partitioned into two sets of functions with distinct type signatures --- local rules $\mathcal{R}^{L}$, which update beliefs, and meta rules $\mathcal{R}^{M}$, which update rules:
{
\begin{align*}
    &\mathcal{R}^{L}_t \coloneqq \{r \in \mathcal{R}_t\ |\ r:\mathbf{B}\times \mathbf{E} \to \mathbf{B} \} \\
    &\mathcal{R}^{M}_t \coloneqq \{r \in \mathcal{R}_t\ |\ r:\mathbf{R} \times \mathbf{B} \times \mathbf{E} \to \mathbf{R} \} 
\end{align*} }%
where $\mathcal{R}^L_t \cap \mathcal{R}^M_t = \emptyset \text{ and } \mathcal{R}^L_t \cup \mathcal{R}^M_t = \mathcal{R}_t$. The rule set may 
include \textit{identity rules}, which trivially return the rules or beliefs from time $t$ at time $t+1$:
{
\begin{align*}
    I^M &\in \mathcal{R}^M_1 \text{ such that } I^M(\mathcal{R}, \mathcal{B}, \mathcal{E}) = \mathcal{R} \text{ and } & \\
    I^L &\in \mathcal{R}^L_1 \text{ such that } I^L(\mathcal{B}, \mathcal{E}) = \mathcal{B} 
\end{align*} }%
for any $(\mathcal{R}, \mathcal{B}, \mathcal{E}) \in \mathbf{S}$.  State updates $\mathcal{S}_{t-1} \to \mathcal{S}_{t}$ are defined by the receipt of new evidence $\mathcal{E}_{t}$, if any, followed by a sequence of two\footnote{Multiple belief updates in immediate sequence can be implemented by setting the corresponding rule updates to the identity.
} rule applications:
{
\begin{align*}
    &\mathcal{B}_{t} = r^L(\mathcal{B}_{t-1}, \mathcal{E}_{t}) \text{ for some } r^L \in \mathcal{R}^L_{t-1} \\
    &\mathcal{R}_{t} = r^M(\mathcal{R}_{t-1}, \mathcal{B}_{t}, \mathcal{E}_{t}) \text{ for some } r^M \in \mathcal{R}^M_{t-1} \\
    & \mathcal{S}_{t} \coloneqq \langle\mathcal{R}_{t}, \mathcal{B}_{t}, \mathcal{E}_{t}\rangle.
\end{align*} }%
\end{definition}
\vspace{-1em}
 
In  order to specify a reasoning algorithm (\cref{alg:reasoning}), we introduce the concept of a \textit{rule selector function}. Because these functions do not impact whether or not a process constitutes reasoning, we define them separately in Def. \ref{def:external_components}, as part of the \textit{reasoner's implementation} of a reasoning process. A full implementation may also involve additional components, such as a goal (or ``stopping rule'') and a trace recording historical reasoning steps, as specified in Def. \ref{def:external_components}.

\begin{definition}[Reasoner components]\label{def:external_components}
A reasoner can contain or generate the following elements (among others), which are extrinsic to the reasoning process itself.
{
\begin{align*}
    & \texttt{s}_\texttt{L}: \mathbf{R} \times \mathbf{B} \times \mathbf{E} \to \mathcal{R}^L& \text{Local rule selector.}\\
    & \texttt{s}_\texttt{M} :  \mathbf{R} \times \mathbf{B} \times \mathbf{E} \to \mathcal{R}^M &\text{Meta rule selector.}\\
    & \texttt{s}_\texttt{stop}: \mathbf{S} \to \{0,1\}& \text{Stopping rule.}\\
    & \texttt{tr}: \mathbf{S} \times \mathcal{R}^{L} \times \mathcal{R}^{M} \times \mathbf{S} \to \Sigma^* & \text{Trace writer.}\\
    &\mathcal{T} \coloneqq \left\{\texttt{tr}\left(\mathcal{S}_{i-1}, r^L_i, r^M_i, \mathcal{S}_i\right)\right\}_{i=1}^{T}& \text{Reasoning trace.} \\
    \intertext{$$\text{where } r_i^L \coloneqq \texttt{s}_\texttt{L}(\mathcal{R}_{t}, \mathcal{B}_{t}, \mathcal{E}_{t+1}) \text{ and } r_i^M \coloneqq \texttt{s}_\texttt{M}(\mathcal{R}_{t}, \mathcal{B}_{t}, \mathcal{E}_{t+1}).$$}
\end{align*}
}%
\end{definition}
\vspace{-4em}

\textit{Rule selectors} use the current state's rules and beliefs along with any new evidence, and output a single rule. The black-box nature of the rule selectors in Def. \ref{def:external_components} is powerful: the freedom to implement selectors in any way (hard-coding, learning from data, or hybrid) is the bridge between symbolic and ML interpretations of reasoning. The \textit{stopping rule} uses the current state to output a boolean expressing whether or not to end the reasoning process. Often, the stopping rule will encode the end-goal of reasoning, evoking the goal-directed nature of a reasoner under Def. \ref{def:reasoner}. The \textit{trace writer} considers the selected rules and resulting state change, and optionally outputs a string (using alphabet $\Sigma$) to include in the \textit{reasoning trace}.

With these definitions in place, we describe a generalized \textit{reasoning algorithm} in \cref{alg:reasoning}.

\newcommand{\pluseq}{\mathrel{+}=}

\setlength{\textfloatsep}{5pt}
\begin{algorithm}[t]
    \begin{algorithmic} 
    \footnotesize
    \caption{\footnotesize Valid reasoning as exact rule application.}\label{alg:reasoning}
    \STATE \textbf{Input.} Initial rules $\mathcal{R}_0$, beliefs $\mathcal{B}_{0}$, evidence stream $\{\mathcal{E}_i\}_{i=0}^T$, stopping rule $\texttt{s}_\texttt{stop}$. \\ \vspace{2mm}
    % \jm{Maybe we want some notion of a query here, such that the stopping rule is ``query answered''}
    %%
        %\STATE $\mathcal{S}_0 \gets \langle \mathcal{B}_{0}, \mathcal{E}_0, \mathcal{R}_{0} \rangle$ %\COMMENT{Initialize world state.}
        %%
        %\FOR {$t \in [1, ... , T]$} 
        \STATE $\mathcal{R}, \mathcal{B}, \mathcal{E} \gets \mathcal{R}_{0}, \mathcal{B}_{0}, \mathcal{E}_{0}$ 
        \STATE $\mathcal{S}\gets (\mathcal{R}, \mathcal{B}, \mathcal{E})$%(\mathcal{R}_0, \mathcal{B}_0, \emptyset)$
        \STATE $t \gets 0$
        \WHILE{\textbf{not }$\texttt{s}_\texttt{stop}(\mathcal{S})$}
        \STATE $\mathcal{E}' \gets \mathcal{E}_{t+1}$ \vspace{1ex}
        \STATE $r^L \gets \texttt{s}_\texttt{L}(\mathcal{R}, \mathcal{B}, \mathcal{E}')$ \COMMENT{Select local rule.}
        \STATE $\mathcal{B}' \gets r^L(\mathcal{B}, \mathcal{E}')$  \COMMENT{Apply local rule, update beliefs.} \vspace{1ex}
        \STATE $r^M \gets \texttt{s}_\texttt{M}(\mathcal{R}, \mathcal{B}', \mathcal{E}')$ \COMMENT{Select meta rule.}
        \STATE $\mathcal{R}' \gets r^M(\mathcal{R}, \mathcal{B}', \mathcal{E}')$ \COMMENT{Apply meta rule, update rules.}\vspace{1ex}
        \STATE $\mathcal{S}' \gets (\mathcal{R}', \mathcal{B}', \mathcal{E}')$
        \STATE $\mathcal{T}.\texttt{append}(\texttt{tr}(\mathcal{S}, r^L, r^M, \mathcal{S}'))$ \COMMENT{Update trace.}\vspace{1ex}
        \STATE $\mathcal{R}, \mathcal{B}, \mathcal{E}, \mathcal{S} \gets \mathcal{R}', \mathcal{B}', \mathcal{E}', \mathcal{S}'$
        \STATE $t \pluseq 1$
        %\COMMENT{Update prior rules using new evidence.}
        %%
        %\STATE $\mathcal{B}_{t} \gets \mathcal{R}_t(\mathcal{B}_{t-1}, \mathcal{E}_t)$ %\COMMENT{Apply new rules to prior beliefs and new evidence.}
        %%
        %\STATE $\mathcal{S}_t \leftarrow \langle \mathcal{B}_{t}, \mathcal{E}_t, \mathcal{R}_{t} \rangle$ %\COMMENT{Update world state.}
        %%
        \ENDWHILE
        %\ENDFOR
        \STATE Return $\mathcal{B}, \mathcal{T}$ % \subseteq \{\mathbf{S}\}_{t=1}^T$. Subset of historical world state.
    \end{algorithmic}
\end{algorithm}

\begin{remark}[How is Definition  \ref{def:reasoning_formal} operational?]
    Operationalization of Def \ref{def:reasoning_formal} and Algorithm \ref{alg:reasoning} is illustrated by implementing a trivial logical reasoning process in Python, offering a simple demonstration for how each core component can be instantiated in compilable code and how validity can be auditable.\footnote{\href{https://github.com/jmaasch/valid\_reasoning}{https://github.com/jmaasch/valid\_reasoning}} 
    In general, the onus is on the researcher to map the core components of Def \ref{def:reasoning_formal} to their unique problem setting, justify the absence of any components, and confirm that validity is present.
    We provide a checklist of best practices in Appendix \ref{sec:checklist}.
\end{remark}
    \vspace{-1em}

\subsection{Examples from Domain-Specific Reasoning}        
\label{sec:domain_specific_reasoning}

Defs. \ref{def:reasoning_informal} and \ref{def:reasoning_formal} are intentionally broad, and indeed a large number of phenomena could be said to satisfy them. To illustrate their flexibility, we map them to specific forms of reasoning that are commonly encountered in mathematics, computer science, and AI. We consider these specific forms of reasoning to be special cases of Def. \ref{def:reasoning_formal} that vary in how rules, beliefs, and evidence are defined or obtained. See Appendix \ref{appendix:additional-examples} for additional examples. Table \ref{tab:reasoning-examples} compares all examples by the nature of rules, beliefs, and evidence. For strong examples of operational definitions for reasoning in mathematics, see Defs. 1 and 2 in \citet{zhang2025dagmath}.

\begin{example}[Logical deduction] \label{example:logical_deduction}
    Our framework is  heavily inspired by deductive systems (e.g., Hilbert systems, sequent calculi, natural deduction, or resolution calculi) over classical first-order logic, although it is not limited to these  settings.\footnote{Deductive systems encompass proof systems and formal semantics for zeroth, first, and higher-order logics, and additionally form the basis for automated theorem provers, SMT solvers, and proof assistants; each of which satisfy Def. \ref{def:reasoning_formal}.} Concretely, a \textit{natural deductive system} over a formal language is initialized with a set of {\color{beliefs} premises} $\Gamma$, and a static set of {\color{rules}inference rules} (e.g., \textit{modus ponens} or \textit{modus tollens}) acting on premises. A {\color{process}derivation (\textit{deduction})} of a {\color{conclusions}conclusion} $\varphi$ is a {\color{process}finite sequence of premises} where each is either in $\Gamma$, or obtained from {\color{beliefs}earlier formulas in the sequence} by application of an {\color{rules}inference rule}. If such a derivation exists, $\varphi$ %is said to 
    satisfies the consequence relation $\Gamma \vdash \varphi$. Derivations yield a  {\color{novel}  monotonically increasing belief set} in the closure of $\Gamma$ under the logical consequence relation.
    
\end{example}

We note that natural deductive systems are a highly restricted instantiation of Def. \ref{def:reasoning_formal}, such that no {\color{evidence}new evidence} is provided ($\mathcal{E}_i = \emptyset\ \forall\ i$), and the set of inference rules is fixed ($\mathcal{R}^M_i = \{I^L\}\ \forall\ i$). Logical systems other than classical first-order logic can also be expressed under Def. \ref{def:reasoning_formal}; see, for example, nonmonotonic logic in Appendix \ref{example:nonmonotonic}, which allows for principled belief retraction.

\begin{example}[Bayesian inference] \label{example:bayesian_inference}
    Bayesian inference provides principled means of revising beliefs in hypotheses as new evidence emerges. We {\color{process}iteratively refine}  {\color{conclusions}posterior estimate}  $p(\theta \mid \mathcal{D})$ for unknown parameters $\theta$ by repeatedly applying {\color{rules}Bayes' rule} (Equation \ref{eq:bayes_rule}) as our {\color{beliefs}prior} over $\theta$ and {\color{evidence}observed data} $\mathcal{D}$ {\color{novel}update across time $t$}:
    \begin{align}
        {\color{rules}r_{bayes}} &\coloneqq  
        \left \{ {\color{conclusions}p(\theta \mid \mathcal{D})} = \frac{{\color{evidence}p(\mathcal{D} \mid \theta)} \; {\color{beliefs}p(\theta)}}{{\color{evidence}p(\mathcal{D})}}  \right \} \label{eq:bayes_rule}.
    \end{align}
\end{example}
\vspace{-1em}
Conclusion $p(\theta \mid \mathcal{D})$ is always valid when  $r_{bayes}$ is applied, though it might be biased with respect to ground truth. 
\begin{example}[Reinforcement learning] 
\label{example:rl}
    RL is the ML paradigm concerned with training optimal {\color{goal}goal-directed} decision-makers (i.e., \textit{agents}) through {\color{process}sequential interactions} with an {\color{novel}uncertain environment}. The agent learns a \textit{policy} that maps states to actions. Thus, the RL agent meets our informal definition of a goal-oriented reasoner (Def. \ref{def:reasoner}). Update rules in RL often take the following form (\citealt{sutton1998reinforcement}, p. 37):
    \begin{align}
        {\color{rules}r_{update}} \coloneqq {\color{conclusions}\varphi_{new}} \gets {\color{beliefs}\varphi_{old}} + \alpha ({\color{evidence}\tau} - {\color{beliefs}\varphi_{old}})
    \end{align}
    where $\varphi$ is some estimate, $\alpha$ is step size, $\tau$ is the target or a desirable (yet perhaps noisy) direction (e.g., the reward), and $(\tau - \varphi_{old})$ is an estimation error. For example, we can estimate the agent's reward for some action at step $t+1$ as
    \begin{align}
        {\color{conclusions}Q_{t+1}} = \frac{1}{t} \sum_{i=1}^t {\color{evidence}R_i} = {\color{beliefs}Q_t} + \frac{1}{t} [{\color{evidence}R_t} - {\color{beliefs}Q_t}]
    \end{align}
    where $Q_t$ is the estimated $t^{th}$ reward (prior belief)  and $R_t$ is the observed $t^{th}$ reward (evidence).
     
\end{example}

\subsection{Validity \& Soundness}
\label{sec:validity}

We now define \textit{valid} and  \textit{sound}
reasoning. We use the terms validity and soundness as classically used to evaluate logical arguments \citep{copi2016introduction,gensler2017introduction,sep2026logical}. Validity is meant to replace our heuristic use of \textit{true} reasoning with a more concrete concept: any superficially reasoning-like behavior that does not satisfy Def. \ref{def:validity} \textit{is not reasoning}, though it may be useful reasoning emulation.  

\begin{definition}[Validity]\label{def:validity}
    A transition from state $\mathcal{S}_t$ to $\mathcal{S}_{t+1}$ is \textit{valid} if and only if it arises from the application of a rule $r \in \mathcal{R}_t$ to components of state $\mathcal{S}_t$. %a prior state $\mathcal{S}_{t-k}$, for $t \geq k \geq 0$.
\end{definition}

%Following from Def. \ref{def:validity}, we claim the following. 

\begin{claim}[Valid reasoning arises from exact rule application] \label{claim:validity_exact_rule_application}
Validity requires that each rule is always executed exactly: not partially, not approximately, not sometimes. This does not preclude rule-based means of handling stochasticity, uncertainty, and approximate inference.
\end{claim}

We can use Def. \ref{def:validity} to further clarify our definition of an \textit{r}-zombie: a system that generates reasoning-like output but lacks the mechanisms necessary for validity. Note that Claim \ref{claim:validity_exact_rule_application} holds regardless of whether the rule set is observable by the human user. Claim \ref{claim:validity_exact_rule_application} is in line with  treatments in symbolic AI, as well as recent work in generative AI: \citet{zhang2025dagmath} claim that  ``operations must be \textbf{exact}'' in LRM reasoning (original emphasis). %, while \citet{györgy2025exactlearning} argue for a shift away from statistical learning toward \textit{exact learning} to address LRM ``jagged intelligence'' \citep{grand2025self}.  
See \cref{objection:scale} and Appendix \ref{sec:historical} for further discussion of the history and revival of rule-based reasoning.

Unlike valid reasoning, \textit{soundness} requires a notion of correctness or alignment with respect to external assessments.

\begin{definition}[Soundness] \label{def:soundness}
    A valid transition from state $\mathcal{S}_t$ to $\mathcal{S}_{t+1}$ is \textit{sound} if and only if all premises (as encoded by $\mathcal{B}$, $\mathcal{R}$, and $\mathcal{E}$)
    are true with respect to external evaluation. 
\end{definition}

While sound reasoning is always valid, valid reasoning need not be sound \citep{copi2016introduction,gensler2017introduction,sep2026logical}. This gives way to Claim \ref{claim:valididty_independent}.

\begin{claim}[Validity is independent of rule selection] \label{claim:valididty_independent}
    Implementing a reasoning process requires selecting which specific rule to apply at each step. Because validity is independent of soundness, and any properly-typed rule application creates a valid output, the validity of a reasoning process is independent of the algorithm used to select the rule sequence, regardless of external ground truth.
 \end{claim}

Claim \ref{claim:valididty_independent} echoes Broome's (\citeyear{broome2013rationality}) \textit{correctness-by-permissibility}: ``Correct reasoning is not reasoning you are \textit{required} to do by rationality, but reasoning you are \textit{permitted} to do by rationality'' (p. xii; emphasis added). Emphasizing validity over soundness allows for \textit{bounded rationality} in reasoning, where incomplete information and uncertainty can lead the reasoner's conclusions to be ``as much determined by the `inner environment''' (our notion of \textit{state}) ``as by the `outer environment''' (e.g., ground truth) 
 \citep{simon2000bounded}. The import of Claim \ref{claim:valididty_independent} is especially clear when there is no singular objective truth, as it permits  disagreement, subjectivity, and relativism. Crucially, valid reasoning paths do not need to be unique nor reach the same conclusion. Consider pluralism in moral reasoning \citep{snoswell2026beyond}: two moral actors with conflicting moral frameworks could both be said to validly reason even if their  verdicts differ, as long as both exactly apply their respective moral rules. Plurality can also arise in sound reasoning: a single problem often admits multiple sound reasoning paths \citep{wang2023self}, though some paths may be more useful; see \citet{gonzalez2024does} and \citet{maasch2025ccr}, which use commutative diagrams to model this case.

Consequently, Claim \ref{claim:valididty_independent}  highlights that validity says nothing of the optimality, usefulness, nor external correctness of the reasoning process. %Ground truth may even be inaccessible or nonexistent. 
For example, the rule selector could select rules at random, act adversarially, or always return the identity function, and yet the process would still be valid. 

%\rebuttal{When multiple valid reasoning paths lead to the same conclusion, validity does not require identifying the unique correct path — only that the executed path satisfies the rule set and validity conditions operative in that context. In subjective domains, validity conditions may themselves be encoded as rules (e.g., as moral constraints, legal statutes, or cultural priors, formalized as rules per Definition \ref{def:reasoning_formal}). We acknowledge that operationalizing validity in such settings is an open research challenge, and flag it explicitly as a priority for the Benchmark and Evaluations directions in our Calls to Action.*}

%\subsection{Beyond Validity: Usefulness \& Alignment}

\subsection{Additional Implications of Definition \ref{def:reasoning_formal}} 
\label{sec:implications}

\begin{claim}[Reasoning is commonplace]
    The permissiveness of Def. \ref{def:reasoning_formal} may appear to undermine its value, as it admits simplistic and low-utility systems. We argue something different: when distilled to its core components, reasoning is commonplace. The fact that ``reasoning'' admits vacuous and trivial examples, as well as complex phenomena, is a necessary consequence of correctness-by-permissibility. This ordinariness is also evident in human cognition: everyday, we reason for both trivial tasks and complex problem-solving. See Appendix \ref{appendix:contextual_alignment} for further discussion.
\end{claim}

\begin{claim}[A system can be simultaneously an \textit{r}-zombie in one sense and a valid reasoner in another] \label{claim:zombie_and_not}
For example, consider the most rudimentary statistical procedure for next token prediction, denoted $\mathcal{A}$ (Example \ref{example:next_token_prediction}). $\mathcal{A}$ certainly performs probabilistic reasoning over the manifold representing the text in its training distribution. But what if we deploy $\mathcal{A}$ for formal mathematical reasoning? This problem setting requires the sound application of formal mathematical rules at every reasoning step and a deterministic, verifiable numerical output. Now, $\mathcal{A}$ is an \textit{r}-zombie that is misaligned for this deployment context. See Appendix \ref{appendix:contextual_alignment} for further discussion.
\end{claim}

\begin{claim}[Rules are learnable and defeasible in the general case] \label{claim:rules_learnable}
We contend that rule-based reasoning and data-driven ML (e.g., probabilistic deep learning) are not mutually exclusive. \textit{Learnable rules} are essential for tying Def. \ref{def:reasoning_formal} to modern AI and the bitter lesson \citep{sutton2019bitter}: rules do not need to be hard-coded by human domain experts, and the future of autonomous reasoning will likely include systems that learn defeasible rules and beliefs on-the-fly. See  \citet{oh2025discovering}, in which an artificial agent autonomously discovered a SOTA RL rule that outperformed human-designed rules. See Appendix \ref{sec:historical} for further historical perspectives.
\end{claim}

\begin{claim}[Rules are explanations] \label{claim:rules_explanations}
    The explainability of a reasoning process lies in the rule set, as rules are the justifications by which each intermediate reasoning step is executed. In this conceptualization, \textit{rules themselves are  explanations} for how the reasoner reached conclusions. By extension, the absence or unobservability of a rule set results in poor explainability. We contrast this notion of rules-as-explanations with CoT, which is neither necessary nor sufficient for explainability (see \S\ref{sec:problem_significance}).
\end{claim}

\begin{claim}[Operationalization facilitates trust]
   A central aspect of trust is the accurate representation of the capabilities or expected behavior of a system \cite{kaur2022trustworthy}. Validity formalizes an expectation found in many common definitions of reasoning (see Appendix \ref{appendix:alternative_definitions} for further discussion). Claims of ``reasoning'' applied to models which fail to meet a minimal bar of validity thus endanger trust. Similarly, claims about ``reasoning'' without a clear operationalization of the term leave validity and soundness unfalsifiable.
\end{claim}

\begin{claim}[Reasoning requires memory] Notions of prior beliefs, evidence, and rules imply the existence of \textit{memory}, as this body of information must be stored and recalled. This does not preclude special cases of \emph{memoryless} or \emph{Markovian} reasoning processes where all information needed at step $t$ is contained in $\mathcal{S}_{t-1}$, 
as these rely on a persistent representation of the immediately preceding state. Several proposals for autonomous machine intelligence \citep{lecun2022path}, AGI \citep{hendrycks2025definition}, and transformer-based LRMs \citep{cheng2026conditional} explicitly emphasize memory or persistent state as a core component of intelligent behavior. 
\end{claim}

\begin{claim}[Natural language is not necessary for reasoning]
\label{claim:language}
    Defs. \ref{def:reasoning_informal} and \ref{def:reasoning_formal} do not imply a necessary role of natural language in AI reasoning. Similarly, \citet{broome2013rationality} does not assume that natural language is necessary for human reasoning.  Evidence from neuroscience suggests that language may not be required for complex symbolic thought \citep{fedorenko2024language}, deductive reasoning \citep{coetzee2022dissociating}, nor  mathematical and logical reasoning \citep{fedorenko2016language}. Increasingly, neural methods explore reasoning in latent space rather than language space  (\citealt{hao2025latent,zhu2025reasoning,wang2025hierarchical}; \textit{inter alia}). 
\end{claim}

\subsection{Rules \& Validity in Neural Networks}

Major outstanding questions surround the nature of rules and validity in black-box neural reasoning, e.g.: Can neural networks learn rules on-the-fly for general reasoning under distribution shift? Can the parameters of a neural network store rules, and if so, how do we locate them? Can rules be added or removed with fine-grained control? While evidence can be construed as model inputs and terminal beliefs as model outputs, what is the nature of intermediate beliefs? If we assume that rules are indeed embedded in the model's parameters, where are the mechanisms ensuring \textit{exact application} of these rules? %Self-consistency limitations and reasoning brittleness suggest that exactness is not natively supported in LLMs, though scaffolding might improve this. 
While conclusively answering these questions is out of scope for this work, preliminary evidence is available and we offer some speculative comments.

Program synthesis with neural induction is a form of rule learning that has proven useful for abstract reasoning in neural networks \citep{chollet2024arc,li2025combining}. As the discovered rules are expressed in code, exact rule application can be outsourced to a compiler. Evolutionary self-improvement loops in program synthesis \citep{pourcel2025self} can be framed as metarules for rule revision. Test-time training procedures \citep{sun2020test,akyurek2024surprising} can be framed as metarules for on-the-fly rule updating under limited data and distribution shift. 

The computational mechanisms underlying LLM behavior has been explored using \textit{mechanistic interpretability} techniques, including concept probing, network decomposition, and circuit discovery \citep{sharkey2025open}. The \textit{circuit hypothesis} posits that meaningful algorithms (i.e., rules or rule sets) can be identified in network parameters \citep{olah2020zoom}. Recent work \citep{shi2024hypothesis} investigates the properties of purported interpretable circuits, such as greater-than \citep{hanna2023does} and induction heads \citep{olsson2022context}. Machine unlearning might one day extend to suppressing or removing prior beliefs, evidence, or rules in unsafe reasoning, though information removal remains weakly defined and does not offer guarantees on model outputs \citep{cooper2025unlearning}. Steering vectors have been used to guide outputs \citep{wu2025axbench}, though their utility for rule revision is not established. A promising recent direction for validity and soundness in AI reasoning combines LLMs with symbolic scaffolding \citep{belle2025future}, including automated reasoning and formal verification components \citep{wu2024lemur}.

\section{Alternative Views}
\label{sec:alternatives}

See Appendix \ref{appendix:alternative_definitions} for an extended discussion of alternative definitions for reasoning from diverse domains. Here, we comment on mainstream objections to our core theses. %While Objection \ref{objection:scale} argues that \textit{this perspective won't work}, Objection \ref{objection:waste_of_time} argues that  \textit{this perspective isn't needed}.

\begin{objection}\textit{(1) Def. \ref{def:reasoning_formal} violates the bitter lesson \cite{sutton2019bitter}, (2) symbolic AI has already failed, and (3) scaling is all you need}. \label{objection:scale}
We observe several variations of these arguments about rule-based systems, which rightfully highlight the knowledge acquisition bottlenecks and lack of generalization in classical expert systems. %%%

\textbf{Rebuttal: Points (1) and (2) are false, and (3) is speculative.} We acknowledge the historical context of an ``AI winter'' following the ``first wave'' of AI, in contrast to the groundbreaking successes of AI's ``second wave'' (\citealt{Fouse_Cross_Lapin_2020}; Appendix \ref{sec:historical}). We understand that this context may raise skepticism about the feasibility of designing systems that meet our standard of validity. However, we contend that our theses are equally compatible with symbolic  and data-driven methods. Per Claim \ref{claim:rules_learnable}, the learnability of rules makes Def. \ref{def:reasoning_formal} amenable to contemporary ML. Because Def. \ref{def:reasoning_formal} does not require hardcoding nor injection of human domain expertise, it is compatible with the bitter lesson. Though recent advances in generative AI are compelling, outright rejection of symbolic methods is near-sighted: see Lean, a symbolic system for gold-standard automated theorem proving \citep{de2015lean}; the neuro-symbolic AlphaGeometry 2 \citep{chervonyi2025gold} and AlphaProof \citep{hubert2025olympiad}, which can solve Olympiad-level math; recent successes in agentic LLM tool use; \textit{inter alia}. While scaling model parameters, data size, and inference-time compute has resulted in profound performance gains \citep{kaplan2020scaling,bi2024deepseek,muennighoff2025s1}, it remains pure speculation whether scaling is sufficient to reach various goals. Scale has not yet resolved hallucination, explainability, out-of-distribution generalization, or other factors that undermine trustworthy reasoning. A AAAI survey found that 76\% of respondents believed ``scaling up current AI approaches'' was ``unlikely'' to ``very unlikely'' to produce AGI \cite{rossi2025aaai}. 
\end{objection}

\begin{objection} \textit{Empirical performance matters more than theoretical guarantees, so rule-based validity is a waste of time}. \label{objection:waste_of_time}
 We observe a common argument that (1) benchmark accuracy is a sufficient proxy for reasoning and (2) if empirical evaluations yield consistently high scores, then the underlying process is of lesser concern.

\textbf{Rebuttal: Sometimes yes, sometimes no.} Circling back to the discussion in \S\ref{sec:intro}, an essential task in contemporary AI will be thoughtfully  delineating where reasoners are required and where (im)perfect \textit{r}-zombies are sufficient. Relatedly, ``How deep and how reliable does the reasoning have to be in order to do certain important things?'' (Holger Hoos in  \citealt{aaai2025panel}). Indeed, sometimes ``the best is the enemy of the good'' and ``optimizing is the enemy of satisficing'' \citep{simon2000bounded}. However, the fallibility of empirical evaluation becomes especially problematic under distribution shift, rare events, adversarial attack, and safety-critical or high-stakes domains. As discussed in \S\ref{sec:problem_significance}, benchmarks have finite coverage and are prone to design flaws \citep{wallach2025position,alaa2025position,white2025livebench,cheng2025benchmarking}, including the conflation of process (reasoning) with the product of that process (QA accuracy, etc.) \citep{chollet2019measure}. As in formal logic, we hold validity as a prerequisite for soundness (Def. \ref{def:soundness}), so any domain requiring external correctness will require validity guarantees. We contend that many scientifically and economically important use cases require validity, including many decision-making systems with safety or fairness implications (e.g., in medicine, policing, etc.).
\end{objection}

%\rl{Moved commented-out rebuttal / objection 3 to appendix for workshopping - I'm more comfortable with it there.}

\section{Conclusion}

\textbf{Call to Action \;} Based on a synthesis of the literature, we propose an operational definition for rule-based reasoning that is compatible with modern ML. However, this is not the only operational definition that could provide research value. We urge the community to engage with our definitions and claims, identify shortcomings, and propose alternatives. We encourage the application of our scientific communication checklist (Appendix \ref{sec:checklist}) to any reasoning-related research or product communication, with a particular focus on domain-specific operationalization. %We recommend that researchers and marketing professionals \rl{reserve the term ``reasoning'' for processes which meet the standard of validity; and in other cases, to} 
%refrain from using ``reasoning'' to describe processes which fall short of the standard of validity and 
%instead reference formalizations of reasoning emulation (e.g., \textit{r}-zombies). 
We advocate for the prioritization of trustworthiness and auditablity in future research and product releases. Echoing calls for ``interpretability by design'' in  mechanistic interpretability \citep{sharkey2025open},  we strongly encourage researchers to build AI reasoning systems with \textit{validity by design}, particularly in domain-specific settings where validity is legally, ethically, practically, or mathematically mandated. We recommend \ref{thesis:definition} and \ref{thesis:rules} as guiding principles for evaluation design.

\textbf{Limitations \;} We attempt to formalize reasoning in the language of math and pseudocode, as these are actionable for the theoretical and engineering communities that conduct AI research. In particular, they lend themselves uniquely well to operationalization relative to the ambiguities of natural language. Thus, this approach has practical utility for building real systems. However, this work does not offer a formal and extensive philosophical treatment of reasoning, as may be found in analytic philosophy and philosophy of mind. Our proposed definitions are a starting point to spur community engagement with \ref{problem:p1}, \ref{problem:p2}, \textit{r}-zombies, and the other challenges described here. Much work lies ahead for formally and operationally defining reasoning. This will undoubtedly require collaboration among philosophers, mathematicians, and computer scientists. 

\section*{Acknowledgments}
The authors thank Dr. Helen Nissenbaum for edifying discussions on epistemic trust in AI. We thank Dr. Ted Meeds and Dr. Aditya Nori for insightful conversations and feedback during the development of this work. We thank Dr. Jen Semler for guidance on the philosophy of reasoning. Author J. Maasch acknowledges the Cornell Tech Digital Life Initiative Fellowship and the US National Science Foundation Graduate Research Fellowship under Grant No. DGE–2139899.

\section*{Conflict of Interest Disclosure}

The authors do not have any conflicts of interest to disclose.

%\clearpage

\bibliography{references}
\bibliographystyle{icml2026}

%%%%%%%%%%%%%%%%%%%%%%%%%%%%%%%%%%%%%%%%%%%%%%%%%%%%%%%%%%%%

\clearpage

\onecolumn
\appendix

\setcounter{figure}{0} % this has to come before the next command renewals, otherwise appendix links will not work.
\setcounter{table}{0}
\renewcommand\thefigure{\thesection.\arabic{figure}}  
\renewcommand\thetable{\thesection.\arabic{table}}  

\addcontentsline{toc}{section}{Appendix} % Add the appendix text to the document TOC
\part{Appendix} % Start the appendix part
\parttoc % Insert the appendix TOC

%\section{Technical Appendices and Supplementary Material}

%%%%%%%%%%%%%%%%%%%%%%%%%%%%%%%%%%%%%%%%%%%%%%

\section{Checklist: Community Guidelines for Scientific Communication in AI Reasoning Research}
\label{sec:checklist}

\begin{fullchecklist}
\small
\begin{enumerate}[label=\textbf{\arabic*.}]
    \item \textbf{\textit{Definition: Reasoning, Reasoners \& Their Components}}
    \vspace{.5em}
    \begin{enumerate}[label=$\square$\;\textbf{1.\arabic*}]
        \item Reasoning is framed as a process, distinct from any artifact produced by that process.
        \item A formal, operational, and domain-specific definition of reasoning is provided.
        \item Each essential component in Defs. \ref{def:reasoning_informal} and  \ref{def:reasoning_formal} is explicitly defined for the problem setting, where applicable: process, rules, beliefs, evidence, and state. Absence of a component or presence of alternative components is explicitly justified.
        \item Sources of extrinsic evidence are reported.
        \item Research clearly defines the state, if and how it is recorded in memory, and how it is retrieved.
        \item Research clearly reports how reasoning steps are selected, searched for, trialed, etc. If rules are selected by search, the search space and search procedure are defined.
        \item Implementation details for all mechanisms of exact rule application are provided.
        \item Can the system be formally characterized as a goal-directed decision-maker that implements a reasoning procedure (a \textit{reasoner}, Def. \ref{def:reasoner}), or is the system limited to the reasoning procedure itself? 
        \item When a distinct \textit{reasoner} entity is present, its components are clearly described and its goal is operationally defined.
    \end{enumerate}
    
    \vspace{1em}
    \hrule
    \vspace{1em}
    \item \textbf{\textit{Reasoning Process Validity}}
    
    \vspace{.5em}
    \begin{enumerate}[label=$\square$\;\textbf{2.\arabic*}]
        \item Validity is defined w.r.t exact rule application, per Def. \ref{def:validity}. Alternative definitions of validity are rigorously justified.
        \item Conditions for valid transitions $\mathcal{S}_t \to \mathcal{S}_{t+1}$ and exact versus approximate execution are stated.
        \item Is each new belief \textit{provably} obtained by exact rule application, or by some other mechanism? In the absence of proof, hypotheses should be provided.
        \item Research clearly reports the provenance of rules and rule updates.
        \begin{itemize}
            \item Are rules learned, or axiomatic? 
            \item Are rules defined in collaboration with domain experts? 
            \item Are rules continuously updatable? When meta rules exist, how exactly are rule updates obtained?
        \end{itemize}
        \item Potential sources of error are explained, along with means for identifying and preventing invalid reasoning steps.
        \item All theoretical guarantees on validity are formally proven, including formal bounds on performance. Absence of guarantees is clearly stated and justified, and supported by rigorous empirics.
    \end{enumerate}
    
    \vspace{1em}
    \hrule
    \vspace{1em}
    %\item \textbf{\textit{Measurement}}
    \item \textbf{\textit{Evaluation \& Construct Validity}}
    
    \vspace{.5em}
    \begin{enumerate}[label=$\square$\;\textbf{3.\arabic*}]
        %\item Measurement refrains from over-reliance on QA benchmark performance, especially publicly available and popular benchmarks, which are subject to leakage and recall effects.
        %\item Correctness is measured by accuracy of intermediate reasoning steps, rather than solely Question-Answering or output accuracy.
        \item The construct validity of all evaluation methods is explicitly justified w.r.t. the operational definitions under use.
        \begin{itemize}
            \item If evaluation relies on ``reasoning tasks,'' what exactly constitutes a task? How does it capture reasoning behaviors?
            \item Is soundness w.r.t. some external ground truth or preference relevant in this setting? How is it measured?
            \item Are the validity, soundness, etc., of intermediate reasoning steps verified, and if so, how?
            %\item Are users able to view and verify the reasoning process?
        \end{itemize}
        \item Evaluations must clearly address the distinction between the reasoning \textit{process} (relative to internal generating mechanisms) versus the \textit{artifacts} of that process (e.g., QA outputs). Tasks and metrics must measure \textbf{both} process quality and output quality.
    \end{enumerate}
    
    \vspace{1em}
    \hrule
    \vspace{1em}
    \item \textbf{\textit{Utility, Explainability \& Trustworthiness}}
    
    \vspace{.5em} 
    \begin{enumerate}[label=$\square$\;\textbf{4.\arabic*}]
        \item Uses and limitations of the system are clearly defined.
        \item Potential harms from use or misuse of the system are addressed.
        \item All sources of explainability are described, and any absence of explainability is directly justified.
            \begin{itemize}
                \item What role do reasoning traces play, what form do they take, and how observable are they to the user?
                \item Is the reasoning trace theoretically guaranteed to accurately reflect the model's internal process? If not, what reasonable expectations of faithfulness are possible?
            \end{itemize}
        \item The operational definition of reasoning met by the system matches the intended use case. If it falls short, the alignment discrepancy is clearly and thoroughly communicated.
        \item The system implements a useful, nontrivial reasoning process beyond standard ML inference, where usefulness is contextually defined w.r.t. the deployment setting.
        \item Reported findings refrain from excessive or misleading claims, especially in title, abstract,  public reporting to lay audiences, and marketing.
    \end{enumerate}
\end{enumerate}

\end{fullchecklist}

\twocolumn
%%%%%%%%%%%%%%%%%%%%%%%%%%%%%%%%%%%%%%%%%%%%%%

\section{Domain-Specific Reasoning, Continued}\label{appendix:additional-examples}

    \begin{example}[Nonmonotonic reasoning]\label{example:nonmonotonic}
       In this example, we consider \textit{nonmonotonic} (or \textit{defeasible}) logic, which adds a mechanism for principled belief retraction and revision to classical logic \citep{strasser2001non}.
    
        Nonmonotonic reasoning describes a 
        {\color{process}  process of deduction and revision} which admits both {\color{rules}\textit{strict} (static)} and {\color{rules}\textit{defeasible} (modifiable)  rules}, including {rules governing belief retraction, priority, and conflict-resolution}.
        {\color{evidence} New, non-defeasible information} available at time $t$, as well as observed contradictions within the  {\color{conclusions}current  belief state} may trigger retractions or updates (via rule application) to {\color{beliefs} prior beliefs} and {\color{rules}existing rules}. This results in a {\color{novel} nonmononically updating belief state}, in which a {\color{conclusions} conclusion} $\varphi$ derived at time $t$ may fail to hold at time $t' > t$.
    \end{example}

    Next, we consider a trivial case of reasoning where rules are hard-coded and evidence is the empty set. 
    
    \begin{example}[Hard-coded algorithm]\label{example:hard-coded}
        A hard-coded algorithm, as represented by a finite, deterministic Turing machine $\mathcal{M}$, can be considered as a reasoning process with a static rule set, such that exactly one applicable local rule (and no meta rules) exist for any given state. At every time step $t$ following an {initial instantiation}, $\mathcal{M}$ {\color{process}executes a fixed procedure}: read the {\color{beliefs} tape symbol} under the head%%at step $t-1$
        , consult a {\color{rules} transition table} to determine the single applicable rule given the {\color{conclusions}current state}, then {\color{novel}write a symbol, move left or right, and change state} or {halt}.
        
        As in deductive reasoning, new {\color{evidence}evidence} is not provided during the reasoning process. Furthermore, all rule selectors are trivial, as only a single state transition is valid at any time step, and the {\color{conclusions}conclusion} is always the belief state if and when $\mathcal{M}$ halts.
    \end{example}

Example \ref{example:hard-coded} illustrates that the ability of a system to map to the formal definition of reasoning is not necessarily meaningful in itself. \textit{Useful} reasoning  will usually require some kind of \textit{alignment} with user preferences, resource constraints, requirements on soundness, or other details of the unique problem setting. For example, a hard-coded algorithm that responds to every input query with the answer ``4'' vacuously meets the standard of rigorous rule application, but fails to align with a domain-specific setting where soundness requires accurate answers to arithmetic queries.

\begin{example}[Probabilistic next token prediction] 
\label{example:next_token_prediction}
This example presents a form of autoregressive probabilistic reasoning over natural language. Consider the $n$-gram language model that maximizes the probability $p(w \mid h)$ of token $w$ given the history $h$ of tokens preceding $w$   \citep{jurafsky2025speech}. {\color{rules}Rule set} $\mathcal{R}_t$ encodes assumptions over the number of relevant preceding tokens in $h$, along with formulae for valid estimation.  
For example, we can define $\mathcal{R}_t$ as the set containing 
\begin{align}
    %{\color{rules}\mathcal{R}_t} &\coloneqq
    %\begin{cases}
        p(w_{1:n}) &= \prod_{t=1}^n p(w_t \mid w_{1:t-1})  \label{eq:chain_rule} \\  %& \quad \text{\footnotesize Chain rule of probability.}  \\[1ex]
        p(w_t \mid w_{1:t-1}) &\approx p(w_t \mid w_{t-1}) \label{eq:markov} \\ %& \quad \text{\footnotesize Markov assumption.}  \\[1ex]
        p(w_t \mid w_{t-1}) &= \frac{\mathcal{C}(w_{t-1} w_t)}{\sum_{w'} \mathcal{C}(w_{t-1} w')} \label{eq:mle} \\  %& \quad \text{\footnotesize Maximum likelihood estimation}.  \\[1ex]
        &\approx \frac{\mathcal{C}(w_{t-1},w_t)}{ \mathcal{C}(w_{t-1})} \nonumber \\ %& \quad \text{\footnotesize Per Markov assumption.} \nonumber \\[1ex] 
        \widehat{w}_t &= \argmax_{w_t} p(w_t \mid w_{t-1}) \label{eq:argmax}  %& \quad \text{\footnotesize Most likely next toten.} 
    %\end{cases}
        %%
\end{align}
where Equation \ref{eq:chain_rule} is the chain rule of probability, Equation \ref{eq:markov} is the Markov assumption, Equation \ref{eq:mle} is maximum likelihood estimation and its simplification per the Markov assumption, and Equation \ref{eq:argmax} predicts the most likely next token. Thus, we can compute the maximum likelihood for $p(w \mid h)$ by taking the count $\mathcal{C}$ of $n$-grams beginning with $h$ and terminating with $w$ in the training corpus, normalized by the sum of counts for any $n$-gram beginning with $h$. {\color{process}Iterating the prediction procedure} (Equation \ref{eq:argmax}), we can {\color{novel}extend the length of the output text} one token at a time. Current belief at step $t$ is $\color{conclusions} \widehat{w}_t$ and prior beliefs (intermediate conclusions) are $\color{beliefs}\widehat{w}_{t-1}$, as these are generated by the predictor. The {\color{conclusions}final string $\widehat{w}_{1:n}$} can be framed as the terminal conclusion. The  {\color{evidence}initial token(s)} (or {\color{evidence}\textit{context}}, as in LLMs) can be framed as evidence, as these are extrinsically provided to the predictor. 
\end{example}

Reasoning validity in Example \ref{example:next_token_prediction} arises from the exact application of $\mathcal{R}_t$, which says nothing of soundness (e.g., the factuality of  $\widehat{w}_{1:n}$). It is clear that $\mathcal{R}_t$ is agnostic to factuality, as \textit{truth} is not necessarily high probability (e.g., some factual statements describe extremely rare events, such that their constituent tokens are unlikely to coincide frequently in a text corpus). Even if the final string contains misinformation (as often occurs with hallucinations in LLMs, a more complex instantiation of next token prediction), the probabilistic reasoning expressed in Example \ref{example:next_token_prediction}  would be valid under Def. \ref{def:validity}. The important question is whether this validity rule is \textit{sound} for the desired application. If soundness through factuality were necessary for the end user, 
additional constraints would need to be encoded in $\mathcal{R}_t$.

Table \ref{tab:reasoning-examples} provides a summary of all domain-specific examples presented in this paper. %\rebuttal{Additional contexts, such as specific schema for defeasible logic, approximate inference, and representation transformation, are suggested as topics for future exploration.}
\begin{table*}[h]
    \centering
    \small
    \begin{tabular}{p{2.5cm}p{3.0cm}p{3.2cm}p{3.2cm}p{3.0cm}}
    \toprule
    \textbf{Example} & \textbf{Beliefs $\mathcal{B}_t$} & \textbf{Evidence $\mathcal{E}_t$} & \textbf{Rules $\mathcal{R}_t$} & \textbf{Goal / Conclusion} \\
    \midrule
    Logical deduction &
    Derived formulas in the proof state $\Gamma$. &
    Premises $E_0$ are given at $t=0$ and not updated thereafter. &
    Fixed inference rules (e.g., modus ponens, introduction/elimination rules) &
    Derive a target formula $\varphi$ such that $\Gamma \vdash \varphi$. \\[0.3em]
    \midrule
    Bayesian inference &
    Current posterior $p(\theta \mid D_{1:t})$ &
    Newly observed data $D_t$ (possibly aggregated with prior observations). &
    Bayes’ rule and any auxiliary update rules (e.g., conjugate prior updates, approximation schemes). &
    Obtain updated posterior beliefs $p(\theta \mid D_{1:t})$. \\[0.3em]
    \midrule
    Reinforcement learning &
    Current value function estimates, policy parameters, and internal state representations. &
    Observed environment states, state transitions, and rewards %$(s_t, a_t, r_t, s_{t+1})$ 
    obtained by interaction with the environment. &
    Update rules such as temporal difference or policy gradient updates, plus any meta rules adapting learning rates or architectures. &
    Learn a policy that maximizes expected return (i.e., select approximately optimal actions over time). \\[0.3em]
    \midrule
    Nonmonotonic logic &
    Current set of accepted conclusions, including defeasible ones. &
    New information that may conflict with existing conclusions (e.g., exceptions, defaults). &
    Nonmonotonic inference rules that support belief revision and retraction, plus meta rules for revising the rule set itself. &
    Maintain a coherent, defeasible belief set that updates appropriately under new, possibly contradictory evidence. \\[0.3em]
    \midrule
    Turing machine &
    Current configuration of the machine: tape contents $\sigma_t \in \Sigma^{\ast}$, head position $h_t$, and control state $q_t \in Q$, collectively encoded as $\mathcal{B}_t = \langle \sigma_t, h_t, q_t \rangle$. &
    Input word $w \in \Sigma^{\ast}$ (typically fixed at $t=0$). &
    Transition relation or function $\delta_t : Q \times \Sigma \to Q \times \Sigma \times \{L,R\}$. &
    Compute the value of a (partial) function $f : \Sigma^{\ast} \rightharpoonup \Sigma^{\ast}$ on input $w$, i.e., reach a halting configuration with output tape $\sigma_T$ such that $\sigma_T$ encodes $f(w)$. \\[0.3em]
    \midrule
    Probabilistic next-token prediction &
    Current token given prior tokens. &
    Token(s) provided at initialization (e.g., the first word of a sentence, the prompt to an instruction-tuned LLM, etc.). &
    Probability rules (e.g., chain rule), structural assumptions (e.g., Markovianity), maximum likelihood formulae. & Iterate procedure until query is answered (e.g., string is of desired length, etc.).
    \\[0.3em]
    \bottomrule
    \end{tabular}
    \caption{Example instantiations of $\mathcal{S}_t = \langle \mathcal{B}_t, \mathcal{E}_t, \mathcal{R}_t \rangle$ for the domain-specific reasoning examples in \S\ref{sec:domain_specific_reasoning} and Appendix \ref{appendix:additional-examples}.}
\label{tab:reasoning-examples}
\end{table*}
\\

%%%%%%%%%%%%%%%%%%%%%%%%%%%%%%%%%%%%%%%%%%%%%%

\section{Alternative Views: Extended Discussion}

\subsection{Alternative Definitions for AI Reasoning}
\label{appendix:alternative_definitions}

Given the expansive range of phenomena that could satisfy our working definitions, what phenomena \textit{do not} satisfy Defs. \ref{def:reasoning_informal} and \ref{def:reasoning_formal}?

We review popular alternative viewpoints on what constitutes reasoning here. We discuss whether these alternative definitions are operational and whether they satisfy Defs. \ref{def:reasoning_informal} and \ref{def:reasoning_formal}. Per \ref{problem:p1}, it is not unusual for papers on reasoning to avoid defining reasoning at all. Thus, some of the alternative definitions discussed here are those that we deem to be \textit{implied} by a subset of the literature, if not explicitly stated. While some alternative definitions provided here partially overlap with Defs. \ref{def:reasoning_informal} and \ref{def:reasoning_formal} and may provide research value in some settings, none feature every core component of our operational definitions (per colored highlighting).

We begin with the dictionary. Testament to the hardness of defining latent  constructs like reasoning, even dictionaries can be ambiguous. Consider the self-referential definitions found in Merriam Webster (the oldest and most authoritative American English dictionary), which also conflate reason and another latent construct: intelligence.

\begin{altdef}[Reasoning, \citealt{mw-reasoning}] \label{altdef:merriam_webster}
\textit{Reasoning, noun.} The use of \underline{reason}; the drawing of {\color{conclusions}inferences or conclusions} through the use of \underline{reason}. \\
\textit{Reason, verb.} To use the faculty of \underline{reason} so as to arrive at conclusions; to discover, formulate, or conclude by the use of \underline{reason}; to persuade or influence by the use of \underline{reason}. \\
\textit{Reason, noun.} The power of comprehending, inferring, or thinking especially in orderly rational ways; intelligence; proper exercise of the mind; the sum of intellectual powers.
\end{altdef}

This definition is not operational in multiple senses: how would one measure the ``power of comprehending,'' the ``sum of intellectual powers,'' or ``orderly rational ways''? While this definition frames reasoning as a form of inference (which we do not disagree with), it does not address the substrates on which inference is performed (extrinsic evidence, prior beliefs, etc.) nor any concrete mechanisms by which inference is executed (exact rule application, etc.).

The Stanford Encyclopedia of Philosophy (SEP) does not provide a single authoritative definition, with definitions varying across articles.

\begin{altdef}[Reasoning, \citealt{sep-reasoning-moral} in SEP] \label{altdef:sep_1} ``Active or explicit thinking, in which the reasoner, responsibly guided by her assessments of her reasons \citep{kolodny2005rational} and of any applicable requirements of rationality  \citep{broome2009unity,broome2013rationality}, attempts to reach a well-supported {\color{conclusions}answer} to a well-defined question \citep{hieronymi2013use}.''
\end{altdef}

\begin{altdef}[Automated reasoning, \citealt{sep-reasoning-automated} in SEP]  \label{altdef:sep_2}
    ``Reasoning is the ability to make inferences [by] proving the {\color{conclusions}conclusion} from the given assumptions by the systematic application of {\color{rules}rules} of deduction embedded within the reasoning {\color{process}program}.''
\end{altdef}

Alternative Def. \ref{altdef:sep_1} contains too many ambiguities   to be easily operationalized (``responsibly guided by her assessments'', ``requirements of rationality'', ``well-supported'', etc.). Further, Alternative Def. \ref{altdef:sep_1} invokes Broome's notion of rational \textit{requirement}, which \citet{broome2013rationality} replaced with rational \textit{permissibility} (a stance that we also take in this position; \S\ref{sec:validity}). Alternative Def. \ref{altdef:sep_2} is clearer, and contains some ingredients from operational Def. \ref{def:reasoning_formal}: conclusions drawn by the systematic application of rules as embedded in the reasoning program implies (1) a sequential process of exact rule application and (2) validity, correctness-by-permissibility, etc. Assumptions might encompass evidence, prior beliefs, and/or some forms of rules, though this is unclear. Sources of extrinsic evidence are not directly addressed. This definition also departs from ours in casting reasoning as an \textit{ability} rather than a process. 

Our informal Def. \ref{def:reasoning_informal} closely resembles the definition proposed by \citet{wang2025hierarchical}:

\begin{altdef}[Reasoning, \citealt{wang2025hierarchical}] \label{altdef:hierarchical_reasoning}
    The {\color{process}process of devising and executing} complex {\color{goal}goal-oriented} {\color{rules}action sequences}.
\end{altdef}

Like Defs. \ref{def:reasoning_informal} and \ref{def:reasoning_formal}, Alternative Def. \ref{altdef:hierarchical_reasoning} frames reasoning as a sequential process. We can map  ``action sequences'' to our concept of rule sequences: both act on evolving streams of intrinsic and/or extrinsic information and result in updates to the state. Defs. \ref{def:reasoning_informal} and \ref{def:reasoning_formal} make this even more explicit: rules \textit{act on} prior beliefs and current evidence, and \textit{output} updated beliefs about the state. We can then map the concept of \textit{devising} action sequences to \textit{selecting} rule sequences. However, Alternative Def. \ref{altdef:hierarchical_reasoning} does not explicitly delineate sources of extrinsic information (evidence), nor define concepts comparable to belief and state. \citet{wang2025hierarchical} depart from Defs. \ref{def:reasoning_informal} and \ref{def:reasoning_formal} by placing goal-orientedness \textit{within} the definition of reasoning. We present an alternative view where reasoning itself has no goal, but may be executed by a goal-directed decision-maker (the \textit{reasoner}, Def. \ref{def:reasoner}). This distinction might or might not have consequences for research. Additionally, \citet{wang2025hierarchical} explicitly invoke complexity (without a concrete threshold for what constitutes \textit{complex}), while our definitions intentionally admit trivial cases.

The following two definitions are also similar to Defs. \ref{def:reasoning_informal} and \ref{def:reasoning_formal}, but (1) are not clearly operational and (2) are overly specialized to human cognition (using anthropocentric language like ``mental process,'' ``thinking,'' etc.), which is of unclear value for designing automated systems.

\begin{altdef}[Reasoning, \citealt{broome2013rationality}] \label{altdef:broome}
    Reasoning is a mental {\color{process}process} in which you operate on the contents of your attitudes, following a {\color{rules}rule}.
\end{altdef}

\begin{altdef}[Reasoning, \citealt{huang2023towards}] \label{altdef:huang}
    Reasoning is the {\color{process}process} of thinking about something in a logical and systematic way, using {\color{evidence}evidence} and past experiences to reach a {\color{conclusions} conclusion} or make a {\color{conclusions} decision}.
\end{altdef}

The ``contents of your attitudes`` in Alternative Def. \ref{altdef:broome} could encompass intrinsic beliefs and/or extrinsic evidence, but this is unclear. Alternative Def. \ref{altdef:huang} notes evidence, but it is unclear how ``past experiences'' differ from evidence (where the latter can, in our conceptualization, be derived from interactions with the environment -- i.e., \textit{experiences}). Unlike Alternative Def. \ref{altdef:broome}, Alternative Def. \ref{altdef:huang} does not explicitly invoke rules (though logical rules may be ambiguously implied by ``a logical and systematic way''). We view these definitions as non-operational, as they leave many questions open: What qualifies as ``logical'' and ``systematic''? Which logic system is being used? And what does it mean to ``operate'' on the ``contents of your attitudes''? Our formal definition makes these notions more concrete.

Note that no components of Defs. \ref{def:reasoning_informal} and \ref{def:reasoning_formal} are explicitly present in the  remaining alternative definitions discussed below (per colored highlighting).

\begin{altdef} \label{altdef:search}
    Reasoning is guided search.
\end{altdef}

\begin{altdef} \label{altdef:planning}
    Reasoning is planning.
\end{altdef}

Defs. \ref{altdef:search} and \ref{altdef:planning} share the same logical fallacy: though search and planning can be framed as reasoning (or as subroutines for reasoning), not all reasoning entails search and planning. Thus, reasoning is not search and planning, though search and planning can be reasoning.\footnote{\textit{All squares are rectangles, but not all rectangles are squares}.}

Contemporary LRMs frequently employ search heuristics that enable exploration or deliberation over the solution space, often with self-evaluation \citep{yao2023tree,xie2023self,grand2025self}. We observe that the performance gains conferred by these  heuristics may contribute toward the conflation of search and reasoning itself. Indeed, the relationship between search and reasoning is significant. Relatedly, a recent AAAI survey found that 44.7\% of respondents agreed that ``reasoning involves a search process'' \citep{rossi2025aaai}. While we agree that search can be an effective means of facilitating reasoning, it is not necessary for reasoning and is not reasoning in and of itself. To clarify, we quote \citet{simon1983search}:

\begin{quote} The same problem-solving algorithm can be viewed, now as search, now as reasoning [...] Consider, for example, a simple
theorem-proving program that works forward from a set of axioms, applying its rules of inference to these to obtain new expressions that can be added to the axiom set. When it finishes tracing a path to a desired theorem, it has succeeded. Clearly it is a search algorithm.
At the same time, the theorem prover is adding, at each step of its search,
new propositions that follow logically from its axioms. It is gradually accumulating a larger and larger collection of deduced propositions. Clearly it is reasoning. [...] The search and constraint metaphors focus upon the process of finding the problem solution, while the reasoning metaphor focuses upon the logical validity of the linkage between initial problem state and solution. Search is centrally concerned with discovery, reasoning with proof.
\end{quote} 

Many principled search procedures can be framed as special cases of Def. \ref{def:reasoning_formal}, and discovery-via-search can be an effective strategy or subroutine for implementing proof-by-reasoning in some settings. However, many counterexamples exist where reasoning does not entail any search process (e.g., Example \ref{example:hard-coded}). Further, search may be present in non-reasoning processes (e.g., search is employed but has no bearing on the final answer, which is obtained by  guessing or memorization). Thus, we conclude that (1) search is not necessary nor sufficient for reasoning; (2) the definition of search does not equate to a general operational definition for reasoning, as accomplished with Def. \ref{def:reasoning_formal}; and (3) we caution against conflating the two in the general case.

\begin{altdef}
    Reasoning is test-time scaling.
\end{altdef}

Scaling test-time compute \citep{snell2025scaling} is a dominant strategy for improving reasoning benchmark performance \citep{bi2024deepseek,chollet2024arc,muennighoff2025s1}.  In this vein, \citet{ye2025emergence} take thinking and reasoning synonymously, defining these as ``the ability to take more time and compute during inference with the goal of producing a higher quality output to a given input.'' This evokes System 2 thinking: the slower, more deliberative, intentional, and logical mode of reflection modeled by Kahneman  (\citeyear{kahneman2011thinking}). Like search and CoT, test-time scaling is a \textit{means of facilitating reasoning} that is nevertheless not necessary for reasoning. See Def. \ref{def:reasoning_formal} (which says nothing of the scale of computational resources and admits trivial implementations) and Example \ref{example:hard-coded} as a counterexample. Additionally, we can imagine an \textit{r}-zombie that adversarially extends its processing time to emulate deliberation or System 2 thinking, without actually engaging in the rule-based mechanisms of valid reasoning. Thus, test-time scaling is not reasoning in itself.

The following two definitions share a common shortcoming.

\begin{altdef} \label{altdef:output}
    Reasoning is correct output.
\end{altdef}

\begin{altdef} \label{altdef:benchmark}
    Reasoning is strong performance on \textit{reasoning tasks} (Def. \ref{def:reasoning_task}): benchmark tasks that would require a human test-taker to perform reasoning.
\end{altdef}

Generative AI papers that target ``strong reasoning performance'' often do not define reasoning \citep{muennighoff2025s1}, inadvertently contributing to the conflation of task accuracy and the reasoning process itself. Our main disagreements with Alternative Defs. \ref{altdef:output} and \ref{altdef:benchmark} are described in \S\ref{sec:problem_significance}. In short, the output-based view (conflating \textit{process} and \textit{product}) and output-based benchmark accuracy are not sufficient for proving that a system engages \textit{mechanisms} of reasoning. Reliance on benchmarks to demonstrate behavior evokes Dijkstra's warning: ``Testing shows the presence, not the absence of bugs'' \citep{dijkstra1970techniques}. Elaborating further, Dijkstra's warning gives way to recommendations on \textit{correctness-by-design}, analogous to our recommendation for \textit{validity-by-design} in reasoning research.

\begin{quote}
    Today a usual technique is to make a program and then to test it. But: program testing can be a very effective way to show the presence of bugs, but is hopelessly inadequate for showing their absence. The only effective way to raise the confidence level of a program significantly is to give a convincing proof of its correctness. But one should not first make the program and then prove its correctness, because then the requirement of providing the proof would only increase the poor programmer’s burden. On the contrary: the programmer should let correctness proof and program grow hand in hand \citep{dijkstra1972humble}. 
\end{quote}

Relying on empirical task evaluation alone is especially fraught when the form of reasoning under study does not feature known or unique ground truth outputs, as in moral reasoning \citep{snoswell2026beyond} or exploratory problem settings (e.g., scientific discovery). We direct the reader to \citet{bowman2021will,cheng2025benchmarking,alaa2025position,weidinger2025toward,wallach2025position,mitchell2025talk} for further reference on the problems associated with benchmarking. 

\subsection{Alternative Views on Rules}

We take a particular stance on rules, framing them as learnable and revisable \textit{operators}, \textit{functions}, or \textit{maps}. However, rules can be otherwise conceptualized.  Wittgenstein's Rule-Following Paradox \citep{kripke1991wittgenstein} concerns the indeterminacy of what rule a speaker is following given any finite set of past behavior. Our framework defines rules as explicit formal objects: functions with defined type signatures (Def. \ref{def:reasoning_components}), rather than norms inferred from behavior. The Rule-Following Paradox applies to fuzzy rule attribution, while our formal definitions concern explicit rule specification and verifiable execution.

Our conceptualization does not assume much in the way of \textit{meaning}, unlike some prior frameworks. The Symbol Grounding Problem \citep{harnad1990symbol} is concerned with how formal symbols acquire meaning. We note that our framework takes no stance on grounding: Def. \ref{def:reasoning_formal} is intentionally agnostic to what the rule set, belief set, and evidence set \textit{mean}, requiring only that rules are applied exactly. 

%However, both problems become relevant when moving from fully-formalized to neural implementations, as systems that learn rules from data face both attribution and grounding challenges. This relevance is further heightened in natural language models. We flag this as a major open research direction.

\section{Extended Discussions}

\label{appendix:discussions}

\subsection{Contextual Alignment }
\label{appendix:contextual_alignment}

The permissiveness of Def. \ref{def:reasoning_formal} may appear to undermine its value, as it admits simplistic and low-utility systems. We argue something different: when distilled to its core components, reasoning is commonplace. The fact that ``reasoning'' admits vacuous and trivial examples, as well as complex phenomena, is a necessary consequence of correctness-by-permissibility (p.7). This ordinariness is also evident in human cognition: everyday, we reason for both trivial tasks and complex problem-solving. In this section, we argue that this fact has important consequences for research, and especially for concepts of alignment \citep{leike2018scalable}.

First, if such a range of computational procedures can be shoehorned into Def. \ref{def:reasoning_formal}, then what useful distinction do \textit{r}-zombies provide? Do \textit{r}-zombies even exist? We contend that many systems meet the standard of valid reasoning only in the most trivial sense. The more interesting and important distinction is whether a system is a \textit{nontrivially aligned reasoner} with respect to deployment context. Does an AI reasoner conform appropriately to domain-specific demands, or is it misadvertised? For instance: can a logical reasoning system perform valid moral reasoning? Can a legal reasoning system perform spatial reasoning? This thought exercise gives way to Claim \ref{claim:zombie_and_not}.

\begin{claim}[A system can be simultaneously an \textit{r}-zombie in one sense and a valid reasoner in another] \label{claim:zombie_and_not}
\end{claim}

For example, consider the most rudimentary statistical procedure for next token prediction, denoted $\mathcal{A}$ (Example \ref{example:next_token_prediction}). $\mathcal{A}$ certainly performs probabilistic reasoning over the manifold representing the text in its training distribution. But what if we deploy $\mathcal{A}$ for formal mathematical reasoning? This problem setting requires the sound application of formal mathematical rules at every reasoning step and a deterministic, verifiable numerical output. Now, $\mathcal{A}$ is an \textit{r}-zombie that is contextually misaligned. This example gives way to our final claim.

\begin{claim}[Useful reasoning provides nontrivial contextual alignment with respect to deployment setting]
% Could also be said of responsible reasoning
    The onus is on the researcher to rigorously justify that the claimed form of reasoning nontrivially satisfies the requirements of the problem setting in which the system is deployed (e.g., soundness, transparency, rule types, evidence sources, etc.).
\end{claim}

Under this argument, we reach an important set of open problems, e.g.: How can we differentiate contextually aligned from trivial and misaligned reasoning, especially in black-box neural models? How can we design contextually aligned autonomous reasoners at scale, especially for settings that require high degrees of transparency, formal verification, or other strict dictates? Answering such questions is an important area for future inquiry.

\subsection{Epistemic Trust in Generative AI \& AI Reasoning}
\label{appendix:trust}

In psychology, trust can be framed as a mechanism for mitigating   uncertainty, reducing resource costs when engaging with external entities, and increasing the probability of successful outcomes  \citep{lukyanenko2022trust}. Science is fundamentally a ``collective epistemic enterprise,'' and as such \textit{epistemic trust} (Def. \ref{def:epistemic_trust}) underpins scientific integrity through two main social contracts: (1) successful collaboration requires that scientists trust the information provided by each other, and (2) societal investment requires that the lay public trusts the information provided by scientists. 

Currently, epistemic trust in AI faces challenges both within the scientific community and with respect to public perception. Reports of public trust vary heavily: 39\% of American respondents predicted that AI will be more beneficial than harmful, versus 83\% of Chinese respondents \citep{maslej2025artificial}; 30\% of Swiss respondents believed AI to be completely unacceptable (up from 23\%), while 26\% supported  human-only decision-making (up from 18\%) \citep{baumann2025reduced}; only 14\% of UK respondents predicted that AI will have a positive impact on society, with negative perception increasing \citep{CDEI2023_DataAI_Wave3}. At the same time, pervasive mistrust coincides with conflicting phenomena: escalating capital investment and user uptake. OpenAI reports 700 million weekly active users for ChatGPT alone \citep{openai2025how}, while 88\% of survey respondents regularly used AI in at least one business function \citep{mckinsey2025state}. %Cognitive offloading, and workforce disruption. 

\ifcomments

\subsection{Historical Perspectives on Reasoning}\label{sec:historical}

\textbf{Philosophy, Logic \& Epistemology \;}  The study of reasoning spans millennia of qualitative and quantitative inquiry. We provide a brief and nonexhaustive summary of historical contributions in the humanities, social sciences, and studies of cognition and the brain.

The history of reasoning is, in many ways, the history of logic and epistemology. Major contributions in ancient logic emanated from early Greek, Indian, Chinese, and Arab cultures, among others. The ancient Greek polymath Aristotle (384--322 BC) provided an early systematic study of logic, establishing deductive reasoning via syllogisms \citep{sep-aristotle-logic}. Aristotle categorized reasoning into \textit{prior analytics} (formal structural argumentation via syllogisms, analogous to notions of validity discussed in this position) and \textit{posterior analytics} (focused on demonstration, definition, scientific knowledge, and inductive reasoning, where premises must be true, primary, immediate, and necessary; this notion maps roughly to soundness and operationalization). We refer the reader to \citet{sep-logic-ancient} for further discussion of ancient traditions of the West.

In India, schools of Buddhist \citep{sep2026buddhist} and Hindu philosophy \citep{britannica2017nyaya} developed rigorous theories of logic and epistemology. Various schemas of inference were proposed for the evaluation of knowledge and arguments (e.g., \textit{premise, reason, example, application, and conclusion}; \citealt{britannica2017nyaya}). In the thirteenth century, the Navya-Ny\={a}ya or Neo-Logical school of Indian philosophy further systematized these logical systems, anticipating aspects of modern set theory and influencing later logicians such as Babbage, Boole, and DeMorgan. See \citet{sep-logic-india} for further discussion of logic in classical Indian philosophy.

The classic epistemological debate over \textit{rationalism versus empiricism} concerns the sources by which we obtain knowledge about our external world \citep{sep-rationalism-empiricism}. While the rationalists emphasized deduction and mathematical certainty (as represented by French polymath René Descartes (1596--1650), German polymath Gottfried Wilhelm Leibniz (1646--1716), et al.), the empiricists  emphasized sensory experience, causation, and probability (as represented by the English philosopher John Locke (1632--1704), Scottish philosopher David Hume (1711--1776), et al.). German philosopher Immanuel Kant (1724--1804) presented a critique of pure reason that attempted to bridge rationalism and empiricism. Modern formal logic (as represented by Gottlob Frege (1848--1925), Bertrand Russell (1872–1970), et al.)  overhauled Aristotelian logic into symbolic mathematical logic. For more recent treatments of reasoning vis-à-vis logic and epistemology in the computer science community (with emphases on probabilistic reasoning and uncertainty), we refer the reader to \citet{fagin1987belief,pearl1990reasoning,fagin1994reasoning,fagin2004reasoning,pearl2014probabilistic,halpern2017reasoning}.

\textbf{Cognitive \& Social Sciences \;} Cognitive science, neuroscience, and psychology have contributed a brain- or mind-centric account of reasoning.
Dual-process theories of reasoning have been explored (and challenged) for centuries \citep{evans2013dual}, perhaps most famously with Daniel Kahneman's theory of \textit{System 1 and System 2 thinking} in psychology and behavioral economics \citep{sloman1996empirical,kahneman2011thinking}. While System 1 is associated with fast, automatic, frequent, and intuitive forms of cognition (e.g., performing basic arithmetic, catching a ball), System 2 entails slow, deliberative, effortful, and logical cognition (e.g., proving a theorem). System 2 thinking is sometimes referenced as a metaphor for inference-time scaling in generative AI. Herbert Simon's theories on \textit{bounded rationality}, reasoning, and decision-making under uncertainty had a significant impact on computer science, economics, and cognitive psychology. As in this position, \citet{simon2000bounded} argues that  interrogating the nature and quality of the \textit{process} of reasoning, and not only its products, clarifies a reasoner's limitations (original emphasis):  
\begin{quote}
    A theory of bounded rationality, then, will be as much concerned with procedural rationality, the quality of the processes of decision, as with substantive rationality, the quality of the outcome. To understand the former, one must have a theory of the psychology of the decision maker; to understand the latter, one needs have only a theory of the goal (the utility function) and the external environment.  [...] When rationality is associated with reasoning \textit{processes}, and not just with its \textit{products}, limits on the abilities of Homo sapiens [sic] to reason cannot be ignored. So the reasoning we find in the classics sounds very different from the calculus of maximization of expected utility in modern neoclassical economics. Taking account of process as well as product is compatible, as neoclassical thinking is not, with the idea that, while human beings usually have reasons for what they do, these are seldom the best reasons, and are seldom consistent over the whole range of their choices.
\end{quote}

\textbf{Automated Reasoning Across ``Three Waves'' of AI \; }  Foundational work on automated reasoning included production systems \cite{davis1984origin,hayes1985rule}, logic programming \cite{lloyd2012foundations}, belief revision \cite{gardenfors1988knowledge, van2008dynamic} and early proof assistants \cite{boyer1975proving,de1983automath, gordon1985hol,coquand1986calculus}, as well as theoretical work on the typed lambda calculus and structural proof theory \cite{howard1980formulae,negri2008structural}. 
The symbolic, rule-based perspective of these approaches dominated early AI research but fell out of favor by the late 1980s, following the collapse of the specialized AI hardware market, unresolved scalability issues in expert systems, and DARPA funding cuts \citep{Fouse_Cross_Lapin_2020}. This period is popularly considered the end of the ``First Wave of AI'' and the beginning of an ``AI Winter'' of reduced global funding and interest in AI. 

In contrast, statistical learning and neural networks drove the fast-paced ``Second Wave of AI,'' as the dominance of deep learning overshadowed rule-based AI through the 2010s \citep{Fouse_Cross_Lapin_2020}. 
Prototypical AI systems from this wave prioritized data-driven approaches, 
viewed models primarily as black boxes, and provided limited explicit reasoning and transparency. Sutton's ``Bitter Lesson'' \citep{sutton2019bitter} was particularly influential in expressing disillusionment with domain-specific understanding in AI, as contrasted with the superior performance of systems relying primarily on scaling laws of increasing compute and training data.

The recent push toward LRMs and renewed interest in formal and neuro-symbolic methods have challenged the perspective that symbolic AI is of mere historical interest  \cite{huang2023towards, belle2025future}. 
Interactive theorem provers such as Lean and Isabelle/HOL \cite{paulson2013proof,de2015lean,blanchette2016hammering} have demonstrated substantial progress toward scalable mathematical formalization and verification. In parallel, the rapid rise of neuro-symbolic architectures in an emerging ``Third Wave of AI'' \cite{garcez2023neurosymbolic} has enabled capabilities such as latent program induction \cite{neelakantan2015neural, macfarlane2025lpn} and theorem-proving systems that tightly integrate symbolic solvers with neural components \cite{xin2024deepseek,chervonyi2025gold}. Further advances in large-scale ML, such as retrieval-augmented generation, model-based planning, and world modeling, have strengthened the case for revisiting classical ideas under modern computational regimes \cite{matsuo2022deep, gao2023retrieval, guan2023leveraging}. 

This shift has been reinforced by growing awareness of the intrinsic limitations of current LLMs (see \S\ref{sec:problem_significance}), including hallucination \cite{xu2024hallucination, bastounis2024consistent}, reliance on heuristics or non-generalizing ``shortcut solutions'' \cite{liu2022transformers,chollet2024arc,xu2025reimagine, mirzadeh2025gsm}, and formal complexity-theoretic boundaries \cite{merrill2022saturated, merrill2023expressive}. 
% As a result, the 2020s have seen a broad push across the field to integrate symbolic reasoning with generative modeling: from chain‑of‑thought prompting and self‑verification pipelines, to structured program‑of‑thought methods, to neuro‑symbolic AGI proposals that aim to combine the flexibility of neural networks with the precision and reliability of rule‑based inference.
We echo  \citet{belle2025future} in hypothesizing that these trends may collectively signal a timely re‑evaluation of rule‑based AI, not as an abandoned ``First Wave'' idea, but as a potential component in next‑generation architectures and in the pursuit of more reliable, transparent, and generalizable reasoning systems.

\section{Glossary}

\begin{definition}[Reasoning task] \label{def:reasoning_task}
    In the AI evaluation setting, we consider a \textit{reasoning task} to be a task that, when previously unseen, would require the average human solver to perform reasoning. Thus, this is an anthropocentric concept that is tied to expectations on human problem solving. 
\end{definition}

\begin{definition}[Operational definition, \citealt{APA_PsychDict_Online}]% APA Dictionary of Psychology\footnote{\href{https://dictionary.apa.org/operational-definition}{https://dictionary.apa.org/operational-definition}}] 
\label{def:operational}
    ``A description of something in terms of the operations (procedures, actions, or processes) by which it could be observed and measured. For example, the operational definition of anxiety could be in terms of a test score, withdrawal from a situation, or activation of the sympathetic nervous system. The process of creating an operational definition is known as \textit{operationalization}.''
\end{definition}

\begin{definition}[Formal verification, \citealt{de2015lean}] \label{def:formal_verification}
    ``Formal verification involves the use of logical and computational methods to establish claims that are expressed in precise mathematical terms. These can include ordinary mathematical theorems, as well as claims that pieces of hardware or software, network protocols, and mechanical and hybrid systems meet their specifications. In practice, there is not a sharp distinction between verifying a piece of mathematics and verifying the correctness of a system: formal verification requires describing hardware and software systems in mathematical terms, at which point establishing claims as to their correctness becomes a form of theorem proving. Conversely, the proof of a mathematical theorem may require a lengthy computation, in which case verifying the truth of the theorem requires verifying that the computation does what it is supposed to do.''
\end{definition}

\begin{definition}[Construct validity, \citealt{sjoberg2022construct}] \label{def:construct_validity}
    A \textit{construct} is a concept that is not directly measurable, but is represented by indicators at the operational level to make it measurable. The validity of a construct (i.e., \textit{construct validity}) is defined by how adequate a concept definition is and how well the indicators represent the concept.
\end{definition}

\begin{definition}[Epistemic trust] \label{def:epistemic_trust}
Per \citet{wilholt2013epistemic}, ``To invest epistemic trust in someone is to trust her in her capacity as provider of information.'' \citet{fonagy2014role} consider epistemic trust to be ``an individual’s willingness to consider new knowledge from another person as trustworthy, generalizable, and relevant to the self.'' Similarly, \citet{irzik2019epistemic} argue that ``Epistemic trust is about taking someone’s testimony that $P$ as a reason to believe that $P$ on the assumption that she is in a position to know whether $P$ and will express her belief truthfully... In the case of scientists, the requirement of good will for epistemic trust amounts to their commitment to the ethical norms of their trade and their sense of obligation to truthfully and accurately share significant knowledge with the public.'' 
\end{definition}

% \begin{definition}[Faithfulness, \citealt{wang2025comprehensivesurveytrustworthinessreasoning}]
%     ``The extent to which [a] model’s outputs align with or are supported by the provided input''. 
% \end{definition}

\end{document}